\documentclass[11pt, a4paper, onecolumn, copyright, goog]{handshaketemplate}

\usepackage{multirow}
\usepackage{tikz}
\usetikzlibrary{calc}
\usepackage{wrapfig}
\usepackage{algorithm}
\usepackage{algpseudocode}
\usepackage{xcolor}
\usepackage{xcolor} 
\usepackage[most]{tcolorbox} 
\usepackage{multirow}
\usepackage[normalem]{ulem}
\usepackage{inter} 
\useunder{\uline}{\ul}{}
\usepackage{enumitem}
\usepackage{rotating}
\usepackage{fontawesome5}
\usepackage{hyperref}

\usepackage[authoryear, sort&compress, round]{natbib}
\usepackage{color-edits}
\usepackage{placeins}
\addauthor[Anjali]{anj}{red}

\definecolor{limeHS}{RGB}{211, 251, 82}
\definecolor{yellowHS}{RGB}{249, 255, 230}
\definecolor{noriHS}{RGB}{5, 35, 38}
\definecolor{cyanHS}{RGB}{122, 243, 255}
\definecolor{darkcyanHS}{RGB}{0, 68, 73}
\definecolor{lightpurpleHS}{RGB}{220, 179, 255}
\definecolor{purpleHS}{RGB}{191, 148, 247}
\definecolor{blueHS}{RGB}{46, 106, 255}

\newcommand{\name}{CUE}

\newtcbox{\highlightbox}[1]{on line, boxsep=0pt, left=2pt, right=2pt, top=2pt, bottom=0pt, colframe=white, colback=#1}

\makeatletter
\DeclareRobustCommand{\pathmark}[2]{%
  \tikz[baseline={([yshift=-0.67ex]m.center)}]
    \node[circle, fill=#1, draw=black, line width=1pt,
          inner sep=0pt, minimum size=1.15em,
          font=\scriptsize\selectfont\bfseries,
          text=black] (m) {#2};%
}
\makeatother

\newcommand{\pA}{\pathmark{limeHS}{A}}
\newcommand{\pB}{\pathmark{yellowHS}{B}}

\newif\ifpreprint
\preprinttrue

\newtcolorbox{fmadialogue}[2][]{breakable, 
    title=#2, fonttitle=\bfseries, fontupper=\fontfamily{Inter-LF}\small\raggedright,  
    colback=white, colframe=purpleHS, 
    boxrule=0.4pt, left=5pt, right=5pt, top=4pt, bottom=4pt, 
    toptitle=1mm, #1}
\newcommand{\fmaturn}[2]{\smallskip\textbf{#1:} #2\par}

\newcommand{\cornerlogo}[2][1]{%
  \begin{tikzpicture}[remember picture,overlay]
    \node[anchor=north west, xshift=2.0cm, yshift=-1.8cm] at (current page.north west)
      {\includegraphics[scale=#1]{#2}};
    \draw[line width=0.3pt]
      ($(current page.north west)+(2.0cm,-2.7cm)$) -- ($(current page.north east)+(-2.0cm,-2.7cm)$);
  \end{tikzpicture}%
}

\title{CUEing User Simulators}
\subtitle{Calibrated User Embeddings for Multi-Turn Benchmarking}

\renewcommand{\today}{}

\author{Anjali Kantharuban$^{1,2}$, Jonas Mueller$^{1}$}
\correspondingauthor{anjaliruban@cmu.edu, jonas.mueller@joinhandshake.com}

\affil{$^{1}$Handshake AI, $^{2}$Carnegie Mellon University}

\begingroup \makeatletter \renewcommand\@makefnmark{}
\makeatother \endgroup

\begin{document}
\begin{abstract}
Recent benchmarks rely on user simulators to evaluate AI agents in multi-turn interaction. 
While existing simulation techniques demonstrate surface \textit{fidelity} to human style and behavior, ecologically valid interactive benchmarking also requires alignment in when and how agents fail across simulated and real user populations.
We find that existing simulators lack \textit{outcome calibration}: agreement with observed success rates and failure patterns when real users interact with the same agent.
We introduce Calibrated User Embeddings (CUE), a framework that both encodes observed sessions and samples continuous representations, then decodes them into persona commands to steer LLMs to act as user simulators without training.
Through this, we evaluate user-conditioned replay of past sessions and aggregate metric agreement when sampling novel personas for the same tasks.   
On $\tau^2$-Bench, CUEd simulators commit fewer simulator-attributed errors and more faithfully reproduce real-user agent failure modes, aggregate success rates, and outcomes for specific task-user pairs than other persona-based simulation methods.
These gains coexist with competitive user fidelity as measured using metrics established in prior work.
After being fit to mostly customer support interactions, the same CUEd simulators generalize to document creation, math tutoring, and casual conversation, and remain effective across different simulator LLMs without CUE retraining.
\end{abstract}

\maketitle
\cornerlogo[0.2]{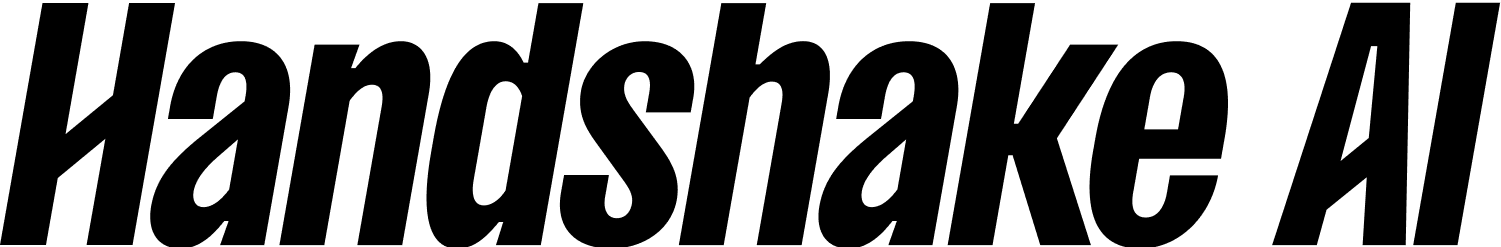}

\begin{center}
\vspace{-1em}
\begin{tabular}{@{}l@{}}
\raisebox{-0.4em}{\includegraphics[height=1.25em]{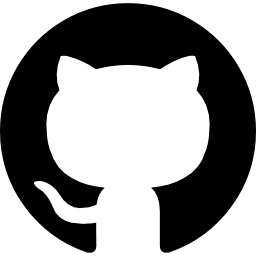}}\ \  \href{https://github.com/Handshake-AI-Research/CUE-user-simulator}{\texttt{Handshake-AI-Research/CUE-user-simulator}} \vspace{0.25em}\\
\raisebox{-0.4em}{\includegraphics[height=1.25em]{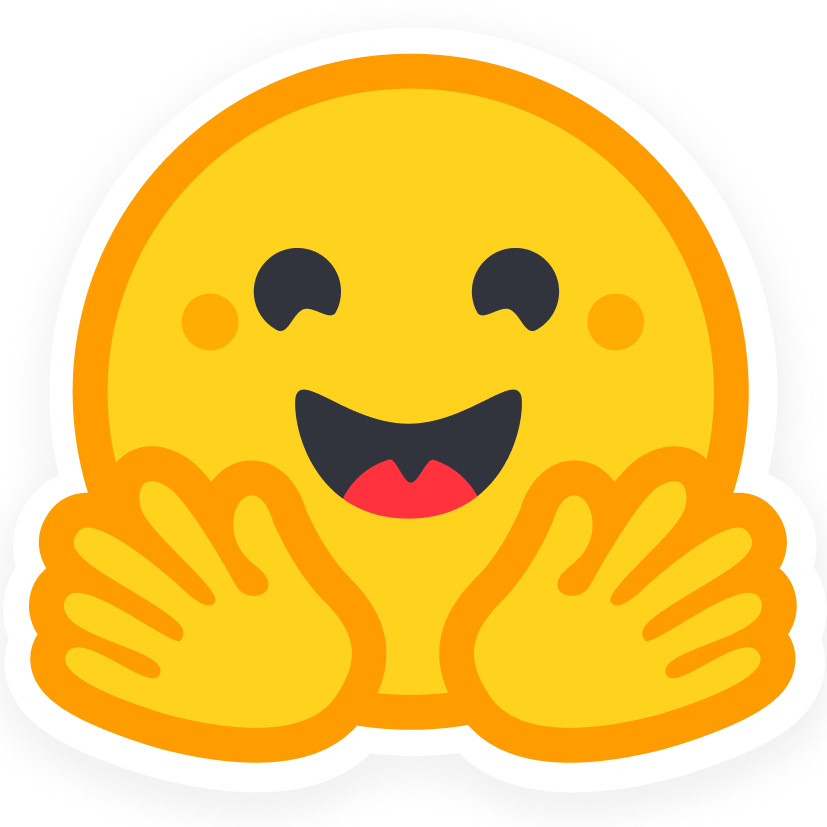}} \ \href{https://huggingface.co/handshake-ai-research/cue}{\texttt{handshake-ai-research/cue}}
\end{tabular}
\end{center}
\section{Introduction}

User simulators are increasingly used to benchmark agents in multi-turn interaction \citep{li2025beyondsingleturn}.
There are multiple axes of assessing the simulators themselves:
\textit{fidelity} asks whether their dialogue resembles human behavior, while
\textit{outcome calibration} asks whether interactions reproduce the agent outcomes observed with human participants.
Calibration here refers to empirical agreement in outcomes, not calibration of predicted probabilities.
A fluent simulator can still abandon a goal, over-cooperate, or fail to elicit agent errors that occur with humans, seen in the divergence in method rankings generated by fidelity and calibration metrics in Figure~\ref{fig:rank}. 
Prior work documents gaps in primarily surface realism \citep{dou2025simulatorarena, seshadri2026lostinsimulation, zhou2026sim2real};
we move beyond fidelity and aggregate success to examine the efficacy of simulators for benchmarking via their performance in role adherence, failure reproduction, and aggregate and user-specific success alignment. 

\ifpreprint
\begin{figure}[t]
    \centering
    \includegraphics[width=\linewidth]{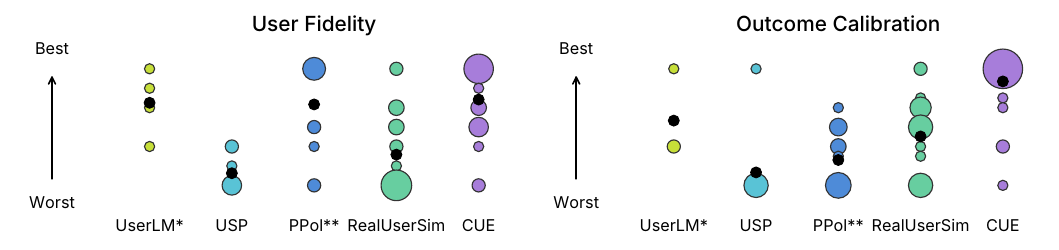}
    \caption{\textbf{Method ranks across user fidelity and outcome calibration.} Each colored point represents a method’s normalized rank for one setting (base model $\times$ evaluation mode), larger points indicate multiple observations at the same rank, and black points indicate mean rank. Ranks are normalized, allowing comparisons with different numbers of applicable methods. \textit{*No user-conditioned replay. **No sampled evaluation.}}
    \label{fig:rank}
\end{figure}
\fi

Existing methods reduce generic simulator artifacts \citep{luo2024duetsim, naous2025flippingthedialogue, chopra2026ppol} or extract profiles from observed conversations \citep{argyle2023outofonemany, zhu2026realusersim, wang2025knowyoufirst}. 
An empirical profile bank already permits repeated simulation, but constrains the available conditioning representations to extracted profiles or their recombination.
A learned continuous representation offers a shared interface for encoding observed sessions and generating additional conditioning vectors.
We examine empirically whether these representations improve behavioral coverage and outcome calibration.

We introduce \textbf{Calibrated User Embeddings (CUE)}, combining a session encoder, a persona-command decoder, training-example retrieval, and a latent diffusion sampler (Figure~\ref{fig:pipeline}).
Our hypothesis is that contrasting human dialogue with generic simulator responses helps suppress simulator artifacts, while contrasts among human sessions retain behavioral differences that diversify interactions. 
The latent representation supports reuse and sampling of this supervision and a text interface makes it usable with black-box LLMs as the base models.
Its name describes the evaluation objective rather than the training loss -- CUE is trained on behavior and style, without benchmark rewards or failure labels in the training objective yet demonstrates improvements in outcome calibration over other persona methods.
Our work makes three contributions:
\begin{enumerate}[nosep]
    \item \textbf{Joint fidelity and outcome evaluation.} We compare conventional fidelity metrics (\S\ref{sec:fidelity}) with success-rate agreement, conditional failure distributions, and retrospective outcome agreement on human-anchored interactions (\S\ref{sec:calibration}).
    \item \textbf{A reusable simulation framework.} CUE  supports either replaying sessions as a particular observed user or simulating new users, demonstrating improved outcome calibration over other persona-based simulators and competitive fidelity (\S\ref{sec:fidelity}, \S\ref{sec:calibration}).
    \item \textbf{Transfer to held-out domains.} A fixed CUE model reduces the writing-outcome error on SimulatorArena and supports fidelity evaluation on SimulatorArena and PRISM across base models (\S\ref{sec:generalization}).
\end{enumerate}
\ifpreprint
    We release our trained CUE model on Hugging Face\footnote{ \url{https://huggingface.co/handshake-ai-research/cue}} and code to reproduce our results on GitHub\footnote{ \url{https://github.com/Handshake-AI-Research/CUE-user-simulator}}.
\fi

\section{Related Work}

\paragraph{Interactive Benchmarks}
As agents increasingly engage in extended user interactions, multi-turn benchmarks have been proposed for conversational ability, task completion, coding, and software interaction \citep{bai2024mtbench, chang2025chatbench, trivedi2024appworld, zhou2024webarena, pan2025benchmarkstalk, qian2025userbench, xu2026theagentcompany, barres2025tau2bench, yang2023intercode}.
Most rely on a user simulator to drive the interaction, but recent work  documents substantial gaps in simulators' realism and aggregate success alignment \citep{dou2025simulatorarena, hu2025simbench, seshadri2026lostinsimulation, suh2026quantifyingutility, zhou2026sim2real}.
However, these works often assume user fidelity -- especially naturalness -- are a strong proxy for outcome calibration when benchmarking.
Empirically, we see that strong user fidelity performance does not necessarily ensure that a simulator is effective for multi-turn benchmarking and outcome calibration requires separate evaluation.

\paragraph{User Simulation}
Existing approaches improve simulation along two main axes.
One line of work adjusts generic LLM simulators to reduce synthesis artifacts such as excessive cooperation, verbosity, formality, or non-ambiguity \citep{luo2024duetsim, vijayvargiya2026ambigswe}.
PPol \citep{chopra2026ppol} evolves persona-generation programs against aggregate realism and coverage objectives.  UserLM \citep{naous2025flippingthedialogue} trains a dedicated, unconditioned user model via LLM fine-tuning.
A second line of work conditions simulators on specific observed users, using explicit or implicit profiles to reproduce individual behavior \citep{argyle2023outofonemany}.
RealUserSim \citep{zhu2026realusersim} extracts structured persona manuals from real conversations. USP \citep{wang2025knowyoufirst} post-trains a simulator to reproduce implicit user profiles.
These methods improve fidelity to users, but user-conditioned methods remain tied to observed human profiles and trained simulators bind the user model to a particular base model, sacrificing the benefits of both scale and rapid model advancement.

\paragraph{Continuous User Representations}
Our work builds on research to learn representations of user-specific linguistic and behavioral variations, including authorship, style, dialect, and preference representations 
\citep{rivera2021luar, wegmann2022sameauthor, kantharuban2026idiolex, patel2025styledistance}. 
Such representations have also been used to condition downstream prediction, recommendation, and generation systems \citep{luo2022federatedrecommendation, zhang2024adpersonalization, liu2025uqabench, ning2025userllm}.
CUE connects this view to user simulation;
instead of representing a user as a discrete profile, command list, or simulator-specific parameterization, it learns a continuous latent representation that can encode an observed user, support generative sampling of additional session representations, and decode into natural-language prompts for steering arbitrary LLMs.

\section{The Calibrated User Embedding (CUE) Framework}

\begin{figure}
    \centering
    \includegraphics[width=\linewidth]{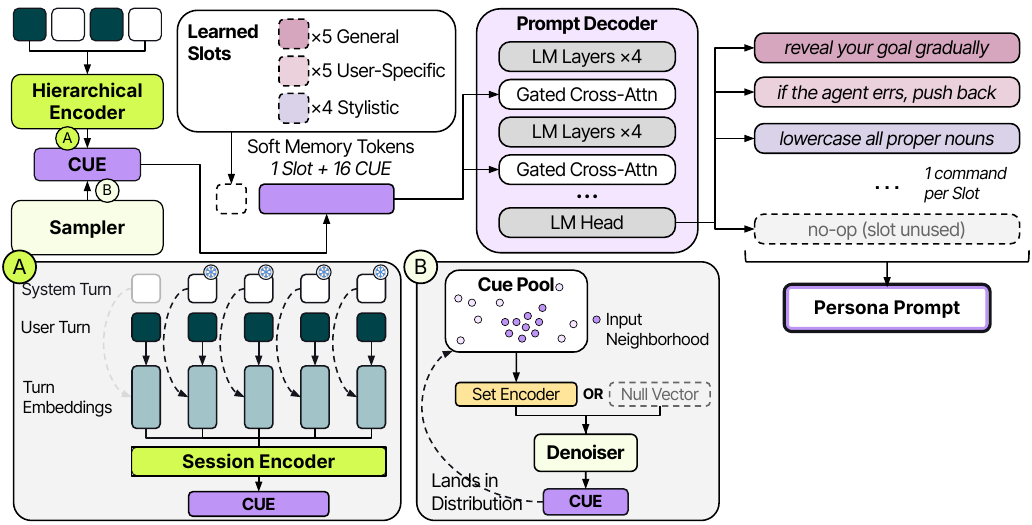}
    \caption{\textbf{The three components of the CUE framework, with two ways to generate CUEs.}
    \protect{\smash{\texorpdfstring{\,\pA}{(A)}}}: Encode an observed session for retrospective replay,
    \protect{\smash{\texorpdfstring{\,\pB}{(B)}}}: Sample a new, distributionally plausible CUE from the learned training distribution. 
    Under either option, decoded commands and retrieved training examples augment the task-specific simulator prompt so the base model may act as the user.
    }
    \label{fig:pipeline}
\end{figure}

Existing user simulators typically represent users as discrete profiles, features, or command lists that can be reused or recombined.
CUE instead learns a continuous representation of the behavior and style expressed in a \textit{session}, supporting both encoding observed interactions and sampling additional representations.
The training pipeline first constructs persona-command targets from dialogue contrasts, then trains a session encoder and command decoder, and finally fits a sampler to encoded training sessions.
At deployment, a CUE is obtained by encoding an observed trajectory or sampling a latent vector (Figure~\ref{fig:pipeline}).
The decoder generates persona commands, while retrieval supplies examples from training sessions.
Together, these augment the benchmark's standard simulator prompt.
The annotation LLM supplies training targets, the learned decoder generates persona commands, and the simulator LLM generates user turns.
CUE integrates these components into a reusable framework for session-conditioned replay and sampled user simulation.

\subsection{User-Conditioning for Replay as Observed Users \texorpdfstring{\,\pA}{(A)}}

Given a dialogue trajectory, the encoder first embeds each user turn while a separate frozen semantic encoder embeds preceding assistant dialogue turns.
Each user-turn representation then attends to the preceding system-turn representation, which has been projected into the user encoder's hidden space.
The contextualized turn embeddings are aggregated into a session-level CUE.
Assistant information therefore enters the CUE through attention, even though the two sides are encoded separately. 
Content-suppression losses reduce, but cannot guarantee removal of, task and outcome information.
To use this CUE with a simulator LLM, a prompt decoder maps the CUE into a set of persona commands.
The CUE is expanded into a fixed soft memory bank and exposed to selected layers via gated cross-attention \citep{alayrac2022flamingo}.
Rather than generating the full persona prompt at once, learned slot embeddings are prepended to the soft memory bank to elicit general behavior, user-specific behavior, or style commands \citep{carion2020end}.
During training, target commands are matched one-to-one to slots within their type, while unmatched slots are supervised toward a no-op target;
at inference, the emitted commands are concatenated into the final persona prompt.
This natural-language interface decouples the learned CUE from any specific simulator LLM.

Training uses a 200-step decoder-format warm-up followed by joint training of encoder and conditioning modules with the decoder LM frozen.
Training combines slot-level command generation with representation objectives designed to preserve user-specific behavior while suppressing task content and maintaining a sampler-friendly latent geometry.
We enforce consistency between original and delexicalized/content-perturbed session CUEs \citep{wu2022infocse}, contrast sessions to avoid collapse towards a mean persona \citep{wu2022infocse}, align CUE similarity with external style similarity \citep{wegmann2022sameauthor}, and regularize variance and covariance across dimensions \citep{bardes2021vicreg}.
Full training details are in Appendix~\ref{app:stage1train}.

\subsection{Sampling Plausible Session Representations \texorpdfstring{\,\pB}{(B)}}

Encoding observed trajectories yields an empirical representation bank.
Sampling additional vectors offers a way to vary conditioning beyond direct bank resampling, without asserting that each vector denotes a distinct person.
We therefore fit a latent diffusion model over the learned CUE distribution \citep{nichol2021improved}.
At inference, it generates a new CUE that can be decoded using the same prompt decoder as observed users.
Unlike direct bank resampling, diffusion generates latent vectors followed by LayerNorm-based normalization;
it does not replace them with nearest bank CUEs.
The sampler may operate unconditionally, over the full training distribution, or condition on a variable-size neighborhood of observed users through a permutation-invariant set encoder over CUEs.
Unconditional sampling targets the learned training-session mixture, while neighborhood conditioning asks the sampler to generate another plausible CUE from a local subpopulation.
Classifier-free guidance supports both modes within a single sampler \citep{ho2022classifier}.
Unless otherwise noted, evaluations use unconditional sampling, with no neighborhood and no classifier-free guidance.
The training mixture is session-weighted and need not match the distribution of people in a target study.
Sampler architecture and training details are in Appendix~\ref{app:sampler}.

\section{Experimental Setup}

\subsection{Training}

We use English multi-turn conversations from 24 DialogStudio datasets \citep{zhang2024dialogstudio}, LMSYS-Chat-1M \citep{zheng2024lmsys}, and WildChat-1M \citep{zhaowildchat}, balancing by capping each dataset at 10k samples.
The mixture includes approximately 170k task-oriented role-play, human-human dialogue, and human-LLM conversations. 
Table~\ref{tab:encdata} lists the source inventory;
only sessions with accepted persona annotations supervise CUE.

For each annotated session, a proposal LLM contrasts the observed user with four simulators' counterfactual replies and other human sessions.
It proposes general behavior, session-specific behavior, and style commands, with evidence-turn references;
lexical and consistency gates filter the targets (Appendix~\ref{app:gates}).
This supervision encourages both correction of simulator artifacts and retention of observed behavioral variation.
Two evaluated base models also occur in the contrast set, while the third (Gemini 3.5 Flash Lite) demonstrates transfer to an unseen family.

The encoder uses a ModernBERT embedding checkpoint, the decoder uses Qwen 3 0.6B and the sampler is an approximately 162M parameter MLP \citep{nussbaum2024nomic, yang2025qwen3, ho2020denoising}.
At inference, greedily decoded commands are supplemented with eight examples retrieved from nearby training sessions (two general and six user-specific).
Thus, the evaluated treatment includes both learned commands and retrieval.
Architecture, training, and generation details appear in Appendix~\ref{app:train}.

\subsection{Evaluation}
\label{sec:evaluation}

\paragraph{Domains}

Our primary evaluation uses 483 human episodes from the $\tau$-USI airline and retail study with $\tau^2$-Bench \citep{zhou2026sim2real, barres2025tau2bench}.
The evaluated agent is GPT 5.2 with high reasoning;
the simulator base model varies independently.
We additionally evaluate on SimulatorArena and PRISM, matching original agents where available and substituting with the same family and size class otherwise.

In \textbf{user-conditioned replay}, CUE encodes the completed human trajectory, including preceding assistant context, and the base model attempts the corresponding task using the resulting persona, with the shared wrapper supplying the task description (Appendix~\ref{app:eval}).
An example of this process for a single task is provided in Appendix~\ref{app:case_study}.
In many practical settings, human interaction data are scarce but available for only a subset of cases or evaluated agent models.
Here, user-conditioned simulation turns those costly observations into reusable, behaviorally grounded replay cases. 
Such replay provides a way to test whether a simulator reproduces the same success/failure patterns before relying on simulation at scale, including for newer agent models for which matched human trajectories are unavailable.
In \textbf{sampled} evaluation, a latent vector sampled without the target human trajectory supplies the persona. 
This tests the learned session mixture, which is an aggregate across many past studies, against the evaluation study population, without assuming that those populations coincide. 

\paragraph{Metrics}

\ifpreprint
    \begin{wraptable}{R}{0pt}
\else 
    \begin{wraptable}[14]{R}{0pt}
\fi
\centering
\footnotesize
\renewcommand{\arraystretch}{1.05}
\begin{tabular}{llll}
\toprule
\textbf{Axis} & \textbf{Metric} & \textbf{Type} & \textbf{Source} \\
\midrule
\textbf{Nat.} & \textbf{S2R-N} $\uparrow$ & Features & \citet{chopra2026ppol}\\
& \textbf{TT} $\downarrow$ & LLM Judge & \citet{wang2026turingrl} \\
\midrule
\textbf{Mim.} & \textbf{AVA} $\uparrow$ & Embedding & \citet{wang2025knowyoufirst} \\
& \textbf{PT3} $\uparrow$ & LLM Judge &  \citet{zhu2026realusersim}\\
\midrule
\textbf{Cov.} & \textbf{S2R-C} $\uparrow$ & Features & \citet{chopra2026ppol} \\
& \textbf{SD-C} $\uparrow$ & Embedding & \citet{patel2025styledistance} \\
\bottomrule
\end{tabular}
\caption{\textbf{User-fidelity metrics adopted from prior work}, by axis. Definitions in Appendix~\ref{app:fidelity_metrics}.}
\label{tab:fidelity_metrics}
\ifpreprint
    \end{wraptable}
\else
    \end{wraptable}
\fi

We evaluate three fidelity axes (Table~\ref{tab:fidelity_metrics}).
\textbf{Naturalness} measures human-like behavior using a behavioral classifier (S2R-N) and a judged pairwise Turing test (TT).
\textbf{Mimicry} measures paired-session similarity using authorship embeddings (AVA) and a five-dimension judge (PT3). 
\textbf{Coverage} measures bidirectional nearest-neighbor proximity using behavioral features (S2R-C) and style embeddings (SD-C).
These are operational proxies established in prior work, described further in Appendix~\ref{app:fidelity_metrics}.

We distinguish outcome calibration from failure attribution. 
\textbf{Success Rate Error $|\Delta|$} ($\downarrow$) is the absolute difference between simulated and human aggregate success rates. 
\textbf{Outcome F1} ($\uparrow$) measures the macro-F1 of environment-assigned binary success on paired human-simulator episodes.
\textbf{Failure Attribution F1} ($\uparrow$) is macro-F1 over agent, user, and environment attribution on paired episodes where both human and simulator fail.
\textbf{Failure Mode F1} ($\uparrow$) is macro-F1 over agent failure modes where both failures are agent attributed.
\textbf{Conditional Failure TVD} ($\downarrow$) compares frequency distributions of agent-attributed error types.
Finally, \textbf{User-Share Gap} ($\downarrow$) is the absolute difference between simulated and human user-attributed shares among labeled failures, reported on the proportion scale.

\paragraph{Baselines}
\label{sec:baselines}

We compare UserLM \citep{naous2025flippingthedialogue}, USP \citep{wang2025knowyoufirst}, PPol \citep{chopra2026ppol}, and RealUserSim \citep{zhu2026realusersim} under a shared harness with matched tasks, agents, and turn budgets.
CUE, USP, and RealUserSim receive paired-session conditioning;
PPol receives the task only, making its mimicry and replay scores task-only references.
UserLM has no individual user conditioning.
We additionally compare with None when possible, which is the base model with no persona injection.
We report sampled coverage for CUE, USP, RealUserSim, UserLM, and None;
PPol's task-conditioned persona generator is omitted from this comparison.
Prompted methods use Llama 3.1 8B, GPT 5.4 Mini, and Gemini 3.5 Flash Lite;
trained baselines use their released base models.

These are end-to-end system comparisons, not matched-supervision controls.
In replay, RealUserSim omits example utterances while CUE retrieves training examples because CUE examples do not come from the same session we are replaying.
Their sampling populations also differ -- CUE's corpus is mixed while RealUserSim is limited to a subset of WildChat.
Appendix~\ref{app:baselines} documents adaptations and Appendix~\ref{app:seed_variance} reports confidence intervals for all metrics for each method.

\begin{table}[t]
\centering
\footnotesize
\begin{tabular}{l|c@{ / }cc@{ / }c|cc|cc}
\toprule
\multicolumn{1}{l|}{\multirow{2}{*}{\textbf{Method}}} & \multicolumn{4}{c|}{\textbf{Naturalness}} & \multicolumn{2}{c|}{\textbf{Mimicry}} & \multicolumn{2}{c}{\textbf{Coverage}} \\
\multicolumn{1}{l|}{} & \multicolumn{2}{c}{S2R-N $\uparrow$} & \multicolumn{2}{c|}{TT $\downarrow$} & AVA $\uparrow$ & \multicolumn{1}{c|}{PT3 $\uparrow$} & S2R-C $\uparrow$ & SD-C $\uparrow$ \\ \midrule
\multicolumn{9}{l}{\textit{Llama 3.1 8B}} \\ \midrule
\multicolumn{1}{l|}{None} & -- & -- & 0.44 & \multicolumn{1}{c|}{0.44} & 0.04 & \multicolumn{1}{c|}{0.19} & 0.01 & 0.45 \\
\multicolumn{1}{l|}{UserLM} & -- & \textbf{0.56} & -- & \multicolumn{1}{c|}{0.41} & -- & \multicolumn{1}{c|}{--} & 0.62 & \textbf{0.55} \\
\multicolumn{1}{l|}{USP} & 0.26 & 0.27 & 0.46 & \multicolumn{1}{c|}{0.47} & 0.06 & \multicolumn{1}{c|}{0.02} & 0.00 & 0.44 \\
\multicolumn{1}{l|}{PPol} & \textbf{0.47} & -- & \textbf{0.18} & \multicolumn{1}{c|}{--} & \textbf{0.26} & \multicolumn{1}{c|}{0.13} & -- & -- \\
\multicolumn{1}{l|}{RealUserSim} & 0.09 & 0.09 & 0.30 & \multicolumn{1}{c|}{0.33} & 0.24 & \multicolumn{1}{c|}{0.17} & 0.00 & 0.46 \\
\multicolumn{1}{l|}{CUE} & 0.41 & 0.48 & 0.22 & \multicolumn{1}{c|}{\textbf{0.26}} & 0.20 & \multicolumn{1}{c|}{\textbf{0.21}} & \textbf{0.63} & \textbf{0.55} \\ \midrule
\multicolumn{9}{l}{\textit{GPT 5.4 Mini}} \\ \midrule
\multicolumn{1}{l|}{None} & -- & -- & 0.44 & \multicolumn{1}{c|}{0.44} & 0.03 & \multicolumn{1}{c|}{0.28} & 0.64 & 0.44 \\
\multicolumn{1}{l|}{PPol} & \textbf{0.67} & -- & \textbf{0.09} & \multicolumn{1}{c|}{--} & \textbf{0.24} & \multicolumn{1}{c|}{0.12} & \multicolumn{1}{l}{--} & \multicolumn{1}{l}{--} \\
\multicolumn{1}{l|}{RealUserSim} & 0.19 & 0.19 & 0.24 & \multicolumn{1}{c|}{\textbf{0.26}} & 0.23 & \multicolumn{1}{c|}{\textbf{0.42}} & 0.55 & 0.57 \\
\multicolumn{1}{l|}{CUE} & 0.53 & \textbf{0.55} & 0.18 & \multicolumn{1}{c|}{\textbf{0.26}} & 0.20 & \multicolumn{1}{c|}{0.33} & \textbf{0.83} & \textbf{0.58} \\ \midrule
\multicolumn{9}{l}{\textit{Gemini 3.5 Flash Lite}} \\ \midrule
\multicolumn{1}{l|}{None} & -- & -- & 0.38 & \multicolumn{1}{c|}{0.38} & 0.08 & \multicolumn{1}{c|}{0.23} & 0.36 & 0.44 \\
\multicolumn{1}{l|}{PPol} & \textbf{0.89} & -- & 0.17 & \multicolumn{1}{c|}{--} & 0.29 & \multicolumn{1}{c|}{0.05} & -- & -- \\
\multicolumn{1}{l|}{RealUserSim} & 0.28 & 0.24 & 0.32 & \multicolumn{1}{c|}{0.34} & \textbf{0.30} & \multicolumn{1}{c|}{0.14} & 0.47 & 0.51 \\
\multicolumn{1}{l|}{CUE} & 0.57 & \textbf{0.59} & \textbf{0.14} & \multicolumn{1}{c|}{\textbf{0.12}} & 0.27 & \multicolumn{1}{c|}{\textbf{0.38}} & \textbf{0.83} & \textbf{0.57} \\ \bottomrule
\end{tabular}
\caption{\textbf{User fidelity on $\tau^2$-Bench.} 
Naturalness entries are user-conditioned/sampled, mimicry uses user-conditioned and coverage uses sampled. Bold marks the best reported value for each variation.}
\label{tab:in_fidelity}
\end{table}

\section{Fidelity to Real Users}
\label{sec:fidelity}

Table~\ref{tab:in_fidelity} reports the user fidelity of various simulators, measured via established criteria from prior work. 
In fidelity, CUE is competitive with current state-of-the-art user simulation techniques.

\paragraph{Naturalness}
PPol leads the S2R-N metric by a large margin but is followed more closely by CUE in their TT values.
The taxonomy used to featurize trajectories for the S2R-N metric is the same as that used to optimize the PPol prompts in their algorithm, meaning it is a biased evaluator for that method specifically.
Omitting PPol for that reason, we see that CUE performs far better on all but one setting than the other baselines for S2R-N.
Relative to the no-persona simulator, CUE improves the TT metric across all six settings, with the no-persona simulator being nearly always distinguishable by the judge.
CUE also improves on RealUserSim's naturalness despite using the same prompt structure, though differences in example access and distribution domain prevent attributing this solely to representation learning. 

\paragraph{Mimicry}
PPol often leads in the AVA metric, whereas the PT3 metric favors other conditioned methods.
Broadly we see that the two metrics disagree often, suggesting that mimicry is difficult to evaluate objectively.
Still, replacing paired CUEs with shuffled CUEs reduces the PT3 value from 0.38 to 0.21 and AVA from 0.27 to 0.20 (Table~\ref{tab:cuemimicry}).
The implementation shuffles episodes within a single domain across tasks, establishing sensitivity to session pairing. 
Thus, the mimicry metrics are sensitive to correct CUE–session pairing, with paired CUEs producing greater similarity to the corresponding observed session than shuffled CUEs, but are not consistent in differentiation between methods.

\paragraph{Coverage}
CUE attains the best or tied-best reported coverage scores, with UserLM also being competitive. 
None reaches a S2R-C value of 0.64 with GPT 5.4 Mini as the base model, but only 0.01 with Llama 3.1 8B;
CUE reaches 0.83 and 0.63 respectively.
As such, we see that the strength of CUE is in the increased variance it encourages in the base model it is applied to.
For Gemini 3.5 Flash Lite, unconditional diffusion sampling performs better in both coverage metrics than resampling from the training session bank used to create the sampler (Table~\ref{tab:cuecoverage}). 
This does not establish that diffusion is necessary for the observed gains, but signals it is one way to diversify outputs.

\section{Calibration to Real Outcomes}
\label{sec:calibration}

\begin{table}[t]
\footnotesize
\centering

\begin{tabular}{lc@{ / }cc|c@{ / }cc|c@{ / }cc}
\toprule
\multicolumn{1}{l|}{\multirow{2}{*}{\textbf{Method}}} & \multicolumn{3}{c|}{\textbf{Role Adherence}} & \multicolumn{3}{c|}{\textbf{Failure Reproduction}} & \multicolumn{3}{c}{\textbf{Success Alignment}} \\
\multicolumn{1}{l|}{} & \multicolumn{2}{c}{User Gap $\downarrow$} & \multicolumn{1}{c|}{\begin{tabular}[c]{@{}c@{}}Attr. \\ F1 $\uparrow$\end{tabular}} & \multicolumn{2}{c}{\begin{tabular}[c]{@{}c@{}}Agent \\ TVD $\downarrow$\end{tabular}} & \multicolumn{1}{c|}{\begin{tabular}[c]{@{}c@{}}Mode \\ F1 $\uparrow$\end{tabular}} & \multicolumn{2}{c}{\begin{tabular}[c]{@{}c@{}} Rate \\ $|\Delta|$ $\downarrow$\end{tabular}} & \begin{tabular}[c]{@{}c@{}}Outcome \\ F1 $\uparrow$\end{tabular} \\ \midrule
\multicolumn{10}{l}{\textit{Llama 3.1 8B}} \\ \midrule
\multicolumn{1}{l|}{None} & 0.14 & 0.14 & \multicolumn{1}{c|}{0.35} & 0.36 & 0.36 & \multicolumn{1}{c|}{\textbf{0.22}} & \textbf{0.21} & \textbf{0.21} & \textbf{0.50} \\
\multicolumn{1}{l|}{UserLM} & -- & 0.54 & \multicolumn{1}{c|}{--} & -- & \textbf{0.31} & \multicolumn{1}{c|}{--} & -- & 0.39 & -- \\
\multicolumn{1}{l|}{USP} & 0.73 & 0.72 & \multicolumn{1}{c|}{0.20} & \textbf{0.31} & 0.93 & \multicolumn{1}{c|}{0.05} & 0.54 & 0.47 & 0.32 \\
\multicolumn{1}{l|}{PPol} & 0.23 & -- & \multicolumn{1}{c|}{0.34} & 0.35 & -- & \multicolumn{1}{c|}{0.13} & 0.45 & -- & 0.38 \\
\multicolumn{1}{l|}{RealUserSim} & 0.29 & 0.27 & \multicolumn{1}{c|}{0.34} & 0.32 & 0.32 & \multicolumn{1}{c|}{\textbf{0.22}} & 0.43 & 0.35 & 0.44 \\
\multicolumn{1}{l|}{CUE} & \textbf{0.13} & \textbf{0.13} & \multicolumn{1}{c|}{\textbf{0.38}} & 0.34 & 0.33 & \multicolumn{1}{c|}{0.20} & 0.35 & 0.32 & 0.46 \\ \midrule
\multicolumn{10}{l}{\textit{GPT 5.4 Mini}} \\ \midrule
\multicolumn{1}{l|}{None} & \textbf{0.08} & \textbf{0.08} & \multicolumn{1}{c|}{\textbf{0.36}} & \textbf{0.24} & 0.24 & \multicolumn{1}{c|}{0.16} & \textbf{0.09} & 0.09 & 0.56 \\
\multicolumn{1}{l|}{PPol} & 0.32 & \multicolumn{1}{l}{--} & \multicolumn{1}{c|}{0.29} & 0.33 & -- & \multicolumn{1}{c|}{0.17} & 0.22 & -- & 0.51 \\
\multicolumn{1}{l|}{RealUserSim} & 0.44 & 0.38 & \multicolumn{1}{c|}{0.31} & 0.33 & 0.27 & \multicolumn{1}{c|}{0.22} & 0.11 & \textbf{0.05} & 0.56 \\
\multicolumn{1}{l|}{CUE} & 0.10 & 0.10 & \multicolumn{1}{c|}{0.31} & 0.25 & \textbf{0.18} & \multicolumn{1}{c|}{\textbf{0.24}} & \textbf{0.09} & 0.10 & \textbf{0.57} \\ \midrule
\multicolumn{10}{l}{\textit{Gemini 3.5 Flash Lite}} \\ \midrule
\multicolumn{1}{l|}{None} & 0.12 & 0.12 & \multicolumn{1}{c|}{\textbf{0.33}} & \textbf{0.35} & 0.35 & \multicolumn{1}{c|}{0.14} & 0.21 & 0.21 & 0.56 \\
\multicolumn{1}{l|}{PPol} & 0.67 & -- & \multicolumn{1}{c|}{0.20} & 0.58 & -- & \multicolumn{1}{c|}{0.09} & 0.34 & -- & 0.42 \\
\multicolumn{1}{l|}{RealUserSim} & 0.62 & 0.63 & \multicolumn{1}{c|}{0.21} & 0.59 & 0.54 & \multicolumn{1}{c|}{0.06} & 0.32 & 0.31 & 0.46 \\
\multicolumn{1}{l|}{CUE} & \textbf{0.06} & \textbf{0.06} & \multicolumn{1}{c|}{\textbf{0.33}} & \textbf{0.35} & \textbf{0.28} & \multicolumn{1}{c|}{\textbf{0.22}} & \textbf{0.12} & \textbf{0.12} & \textbf{0.62} \\ \bottomrule
\end{tabular}

\caption{\textbf{Outcome calibration on $\tau^2$-Bench.} Paired entries are user-conditioned/sampled; F1 metrics are only user-conditioned. Attribution F1 conditions on both episodes failing, mode F1 conditions on both having agent-attributed failures. Bold marks best value per variation.}
\label{tab:failure}
\end{table}

For human-grounded agent evaluation, a simulator should reproduce not only
aggregate success rates but also which types of interactions fail and how.
Table~\ref{tab:failure} compares aggregate outcome agreement with paired reproduction of human outcomes and failure labels.
For $\tau^2$-Bench programmatic rewards identify failed episodes.
We take these and use LLM-assisted thematic analysis to assign an error attribution (user, agent, or environment) and, for agent-attributed failures, a failure mode (Appendix~\ref{app:failure}).
The taxonomy was refined over three researcher-reviewed batches of 100 episodes;
the final batch required only one correction.

\paragraph{Role Adherence}
CUE has the smallest user-share gap among persona-based methods on every base model. Compared with None, CUE has smaller gaps on Llama (0.13 versus 0.14) and Gemini (0.06 versus 0.12), but a larger gap on GPT (0.10 versus 0.08).
These are shares conditional on failure: a change can reflect user-side failures or changes in other failure causes.
Paired attribution macro-F1 adds a case-level check on episodes where both human and simulator fail.
CUE leads on Llama (0.38 vs 0.35 for None), ties on Gemini (0.33), and falls below None on GPT (0.31 vs 0.36).
Broadly, other simulation methods result in poorer alignment of failure attribution despite their fidelity performance, while CUE is in line with the standard simulator used in the benchmark.

\begin{figure}
    \centering
    \includegraphics[width=\linewidth]{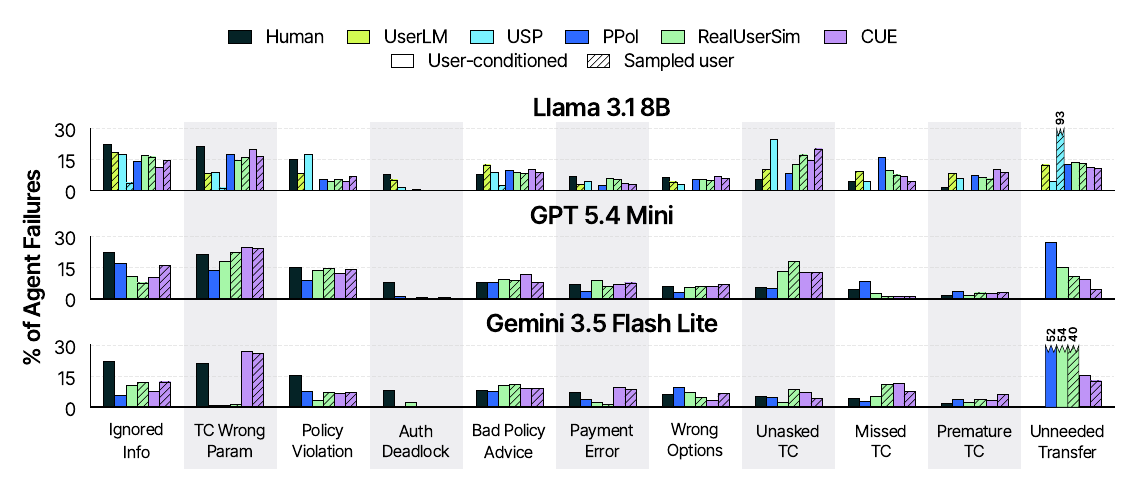}
    \caption{\textbf{Conditional failure composition over agent-attributed failures.} Bars show the proportional distribution of failure modes within each method's agent-attributed failures. \textit{TC = Tool Call}.}
    \label{fig:failure}
\end{figure}

\paragraph{Failure Reproduction}
Figure~\ref{fig:failure} illustrates distortions in failure composition, including authentication deadlocks and unnecessary escalation.
CUE has the lowest sampled agent-failure TVD on GPT (0.18) and Gemini (0.28), while UserLM leads on Llama (0.31), but with little differentiation across all methods except USP. 
User-conditioned failure mode F1 asks whether the specific failure mode agrees when both real-user and corresponding simulated-user episodes have agent-attributed failures.
We emphasize that CUE prompts do not encode task-specific information that explicitly encourages the same failure mode (see Appendix~\ref{app:case_cue_instruction}). Among the evaluated user simulation methods, Table \ref{tab:failure} nonetheless shows that  CUE achieves the highest user-task failure mode macro-F1 when simulating users with GPT or Gemini, and CUE lands within the confidence interval of the highest F1 value when simulating users with Llama.

\paragraph{Success Alignment}
Against the human success rate of 0.61, None is closest on Llama, outperforming all methods.
On the other hand, CUE and RealUserSim close this gap with GPT as the base model and CUE outperforms all other methods including None by a large margin on Gemini.
When examining replay, CUE has the highest outcome macro-F1 among persona methods on all three base models.
Relative to None, however, CUE again displays mixed results with better performance on the two stronger simulator base models. 


\paragraph{Fidelity and Outcome Agreement}
Fidelity does not select a uniformly best outcome simulator -- different metrics rank methods differently.
Although the no-persona baseline often scores below CUE on fidelity, it remains competitive on outcome agreement and outperforms several persona-based methods.
In particular, persona conditioning can increase behavioral resemblance while
simultaneously shifting task difficulty or cooperation in ways that worsen
agreement with human outcomes.
We therefore recommend evaluating outcome calibration directly alongside fidelity when possible rather than assuming that high naturalness, mimicry, or coverage begets alignment for benchmarking.

\section{Transfer to Held-Out Benchmarks}
\label{sec:cueornot}
\label{sec:generalization}

We apply the same CUE model to SimulatorArena, an open-ended cooperative benchmark covering co-writing and tutoring, and PRISM, a demographically diverse casual conversation benchmark, without retraining (Appendix~\ref{app:generalization}).
These are held-out benchmarks with domains outside of task-oriented dialogue.
While CUE's training mixture contains some general conversation from LMSYS and WildChat ($\sim$20k samples), the majority is from task-oriented, so this evaluates whether the system transfers despite the training data imbalance.


\begin{table}[t]
    \centering
        \footnotesize
        \begin{tabular}{@{}ll|cc@{}}
        \toprule
        \multirow{2}{*}{\textbf{Base}} & \multirow{2}{*}{\textbf{Method}} & \multicolumn{2}{c}{\textbf{MAE $\downarrow$}} \\
         &  & Document Quality & Interaction Quality \\ \midrule
        \multirow{3}{*}{{\begin{tabular}[c]{@{}l@{}}Llama\\ 3.1 8B\end{tabular}}} & USP & 5.34 & 1.86 \\
         & RealUserSim & 6.06 & 2.21 \\
         & CUE & \textbf{3.55} & \textbf{1.48} \\ \midrule
        \multirow{2}{*}{{\begin{tabular}[c]{@{}l@{}}GPT 5.4 \\ Mini\end{tabular}}} & RealUserSim & 5.80 & 1.39 \\
         & CUE & \textbf{2.21} & \textbf{1.33} \\ \midrule
        \multirow{2}{*}{{\begin{tabular}[c]{@{}l@{}}Gemini 3.5 \\ Flash Lite\end{tabular}}} & RealUserSim & 4.93 & 1.42 \\
         & CUE & \textbf{2.42} & \textbf{1.41} \\ \midrule
         \multicolumn{2}{@{}l|}{{Human}} & 0.85-1.48 & 0.87-1.50 \\ \bottomrule
        \end{tabular}
        \caption{\textbf{Outcome calibration over SimulatorArena writing benchmark.} Pairwise mean absolute error (MAE) over user-conditioned models using the benchmark's LLM-judge grader.}
        \label{tab:simarenamae}
\end{table}

\paragraph{Writing-Outcome Agreement} 
Table~\ref{tab:simarenamae} reports paired human-simulator mean absolute error on document and interaction quality scores as evaluated by the SimulatorArena benchmark's LLM judge. 
Interaction quality is more akin to user fidelity (examining the surface interaction) while document quality is more akin to outcome calibration (examining the outcome of the interaction).
The scoring is on a scale of 1-10;
for score $s$, $\text{MAE} = \smash{\tfrac{1}{N}\sum_i} |s_i^{\text{sim}} - s_i^{\text{human}}|$. 
The human range provided for context is calculated by performing the same calculation, using two humans who completed the same task as the reference and prediction respectively.
CUE reduces document-quality error for every base model, including 2.21 versus RealUserSim's 5.80 on GPT 5.4 Mini.
Interaction quality differences are smaller (i.e. 1.33 versus 1.39 here) with both CUE and RealUserSim falling within the human range on most base models. 
CUE's document error nevertheless remains above the human comparison range, indicating that two humans performing the same task differ less in document quality than each simulator and its paired human.

\begin{figure}
    \centering
    \includegraphics[width=\linewidth]{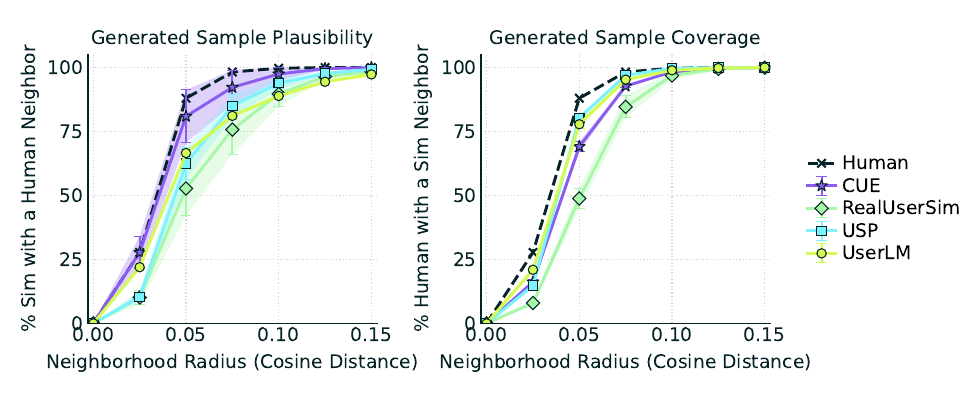}
    \caption{\textbf{Bidirectional stylistic proximity on PRISM.} Left: Percent of generated trajectories with a human neighbor. Right: Percent of human trajectories with a generated neighbor. Methods use the same set for all measurements. Displayed intervals aggregate seeds and backbones into a 95\% CI.}
    \label{fig:prismsweep}
\end{figure}

\paragraph{Bidirectional Stylistic Coverage of Diverse Populations} 
The behavioral coverage scores saturate on PRISM, where all reported methods meet or exceed 0.86.
To examine how well each method covers the user population of PRISM, Figure~\ref{fig:prismsweep} plots neighborhood overlap in the StyleDistance space \citep{patel2025styledistance} using the same 5,284 conversations from both the human study and each method.
The leave-one-out human reference curve demonstrates the density of the human samples, with most samples being within 0.10 of another.
The left panel measures generated-to-human proximity: at a radius 0.05, 81\% of CUE trajectories have a human neighbor compared to 67\% of the nearest alternative.
The right panel measures human-to-generated coverage: at the same radius, USP and UserLM have 80\% and 78\% coverage respectively while CUE has 69\%. 
Notably, generating trajectories close to observed human behavior and covering the observed human population are disparate goals.
At small neighborhood radii, CUE yields the highest generated-to-human proximity, whereas UserLM and USP attain greater human-to-generated coverage.
This illustrates a tradeoff between generating samples near the observed human manifold and covering more of that manifold.

\section{Conclusion}

CUE provides a shared representation and prompt interface for retrospective session replay and sampled user simulation. 
Across $\tau^2$-Bench, SimulatorArena, and PRISM, the complete pipeline is competitive on user fidelity and generally improves outcome agreement relative to other persona-based methods. 
Our results do not establish that the continuous representation or diffusion sampler individually causes these gains, that sampled sessions match the distribution of every target population, or that retrospective replay predicts future user behavior. 
Matched-supervision controls and human studies spanning multiple agents are needed to test these mechanisms and determine whether outcome calibration preserves comparative agent rankings.

Importantly, the strong outcome calibration performance of the no-persona baseline on several $\tau^2$-Bench settings, where it beats several of the baseline methods, shows that adding a more human-like persona does not necessarily produce a better benchmarking proxy.
We therefore recommend evaluating outcome calibration directly rather than treating naturalness, mimicry, or coverage as sufficient evidence of benchmarking validity.

\section*{Limitations}
Our evaluation of outcome calibration rests on two human studies and does not test whether improved outcome reproduction preserved comparative ranking across multiple agents.
The correlation analysis is descriptive because simulator configurations are not independent and the number of method-base model combinations is small.
Our ablations establish that CUEs encode user-specific information and that continuous sampling improves coverage over bank resampling, but we do not isolate every architectural component.
User-conditioned evaluations are retrospective replay experiments and therefore do not establish cross-session or prospective prediction of user behavior.
Additional limitations include the reliance on LLM annotation for failure attribution, containment of all training and evaluation data to English, and the dependence on proprietary simulator base models for some results.
Finally, CUE remains imperfectly calibrated despite performing better than baselines and should not be considered to have resolved all concerns around outcome calibration.

\bibliography{citations}

\newpage

\appendix

\renewcommand{\thefigure}{A\arabic{figure}}
\renewcommand{\thetable}{A\arabic{table}}
\setcounter{figure}{0}
\setcounter{table}{0}

\begin{center}
\huge \textbf{Appendix}
\end{center}
\FloatBarrier

\section{Data}
\label{app:data}

\subsection{CUE Training Data}
\label{app:encdata}

\begin{table}[t]
\centering
\footnotesize
\begin{tabular}{l|rr}
\toprule
\textbf{Dataset} & \multicolumn{1}{c}{\textbf{Train}} & \multicolumn{1}{c}{\textbf{Dev}} \\
\midrule
LMSYS Chat 1M \citep{zheng2024lmsys} & 10,000  & \\
\midrule
WildChat \citep{zhaowildchat} & 10,000 & \\
\midrule
ABCD \citep{chen2021ABCD} & 8,034 &  1,004 \\
AirDialogue \citep{wei2018airdialogue} & 10,000 & 10,000 \\
BiTOD \citep{lin2021BiTOD} & 2,952 & 295 \\
CaSiNo \citep{chawla2021CaSiNo} & 900 &  30 \\
CraigslistBargains \citep{he2018craigslistbargains} & 4,000 &  570 \\
DSTC2 Clean \citep{chao2019DSTC2} & 1,612 & 506 \\
Disambiguation \citep{qian2022disambiguation} & 8,433 & 999 \\
FRAMES \citep{el2017FRAMES} & 1,329  & \\
GECOR \citep{quan2019gecor} & 676 & \\
HDSA-Dialog \citep{chen2019HDSA} & 8,438  & 1,000 \\
KETOD \citep{chen2022ketod} & 4,247  & 545 \\
KVRET \citep{eric2017kvert} & 2,424  & 302 \\
MetaLWOZ \citep{shalyminov2020metalwoz} & 10,000 & \\
MS-DC \citep{li2018mcdc} & 9,985  & \\
MuDoCo \citep{martin2020mudoco} & 6,032 & 690 \\
MultiDoGO \citep{peskov2019muldogo} & 10,000 & 1,150 \\
MultiWOZ 2.1 \citep{eric2019multiwoz21} & 8,434 & 999 \\
MultiWOZ 2.2 \citep{zang2020multiwoz22} & 8,437 & 1,000 \\
SGD \citep{rastogi2020sgd} & 10,000  & 2,482 \\
STAR \citep{mosig2020star} & 6,145  & \\
Taskmaster 1 \citep{byrne2019taskmaster1} & 6,170 & 769 \\
Taskmaster 2 \citep{byrne2019taskmaster1} & 10,000  & \\
Taskmaster 3\citep{byrne2019taskmaster1} & 10,000  & 10,000 \\
WOZ 2.0 \citep{mrksic2017woz2} & 600 & 200 \\
\midrule
\textbf{Total} & \textbf{168,848} & \textbf{32,541} \\
\bottomrule
\end{tabular}
\caption{\textbf{CUE source inventory.} Reported trajectories per source split; related dataset versions are not necessarily unique conversations, and inventory counts are not annotation-retention counts.}
\label{tab:encdata}
\end{table}

The CUE encoder uses the annotated portions of the source datasets in Table~\ref{tab:encdata}. These sources mix naturalistic human-LLM conversations, elicited or role-play dialogue, human-human negotiations, and human-authored or paraphrased task-oriented data. They should not all be described as interactions between real users and deployed agents.
The first two are large, general-purpose dialogue corpora, while the remaining datasets are task-oriented.
Only trajectories for which we construct the supervision described in Section~\ref{app:gates} are used to train CUE; additional unlabeled trajectories are retained as part of the data inventory but are not used for training.

\subsection{Proposal Prompts \& Gating for Training}
\label{app:gates}

For each conversation in the train split, we construct a structured persona manual describing user-specific behavioral and stylistic characteristics.
The labeling procedure uses two complementary contrast sets to distinguish both simulator-like behavior and variation among real users.

At selected user turns, four LLM simulators (Llama 3.1 8B Instruct, Qwen 3 8B, GPT-5.4 Mini, and Claude Haiku 4.5) generate counterfactual responses to the same dialogue context.
These responses provide \emph{simulator contrasts}: examples of how generic simulators differ from the target user.
We additionally retrieve five real-user conversations as \emph{human contrasts}, which help identify behaviors that distinguish the target user from other humans rather than merely from LLM-generated dialogue.

Given the target conversation and these two contrast sets, a proposal model (GPT-5.4 Mini) generates three groups of candidate commands: five general commands derived from simulator contrasts, five user-specific commands derived from human contrasts, and five style commands.
Together, these commands are intended to capture both human-like interaction behavior and variation in how individual users communicate.
Each command is accompanied by evidence turn IDs identifying where the corresponding behavior is expressed in the target conversation.

The proposal prompt explicitly excludes task-specific content, such as identifiers, products, account details, or domain policies, so that commands describe transferable user behavior rather than the particular task being completed.
When behavior depends on dialogue history -- for example, delayed answers, pushback, or escalation -- the prompt requires reactive commands expressed as conditional ``if...then...'' rules rather than unconditional persona attributes.

Candidate manuals are schema-validated, delexicalized, and filtered using deterministic gates:
\begin{itemize}[nosep]
    \item \textbf{Command Coverage.}
    The manual must contain at least 8 commands, including at least 4 simulator-contrast commands and 4 human-contrast commands.

    \item \textbf{Evidence Grounding.}
    Every command must retain at least one valid evidence turn ID after grounding against the target user's turns.

    \item \textbf{Content Leakage.}
    Commands and examples are rejected if they contain task-specific content detected by a fixed lexical filter, including terms associated with credentials, identifiers, products, policies, addresses, receipts, or deliveries.

    \item \textbf{Generic Human Contrast.}
    A manual is rejected if at least 3 human-contrast commands reduce to generic behavioral descriptions such as being polite, cooperative, calm, matter-of-fact, concise, or task-focused.

    \item \textbf{Distinctive Human Contrast.}
    The manual must contain at least 2 human-contrast commands describing behavioral signatures that are present in the target and uncommon among the human negatives.
    These include behaviors such as delayed answering, escalation, pushback, impatience, try-then-report behavior, small talk, playful testing, or acceptance following refusal.
    Generic closers or behaviors shared by most human users do not satisfy this criterion.

    \item \textbf{Emotion Consistency.}
    Commands that attribute frustration, anger, or disappointment are rejected when the transcript contains only acceptance markers (e.g., ``ok'', ``thanks'', or ``that's fine'') and no corresponding negative affect.

    \item \textbf{Answer-Lag Contradiction.}
    We detect answer lag when a user turn responds to an earlier assistant request rather than the immediately preceding one.
    If lag is observed, commands claiming that the user consistently answers the latest prompt in sequence are rejected.

    \item \textbf{Answer-Lag Coverage.}
    If at least 2 answer-lag events are detected, the manual must include a command describing delayed, out-of-order, or previous-question answering.
\end{itemize}

Manuals that fail any gate are regenerated with feedback indicating the failed criteria, for up to three proposal attempts.
The exact lexical filters and gate implementations are provided in the released code.

At rollout time, the resulting persona manual is inserted into the standard simulator prompt for the corresponding domain.
The resulting labeled training set contains approximately 170k examples and constitutes the supervision used to train the CUE models.

\subsection{Evaluation}
\label{app:eval}

Our primary evaluation domain is task-oriented dialogue, but we additionally evaluate on open-ended collaboration and general conversational assistance to test whether simulator quality generalizes across distinct forms of human-AI interaction.

For task-oriented dialogue, we use $\tau^2$-Bench, a customer-service benchmark in which conversational agents interact with users while accessing tools and receiving programmatic rewards \citep{barres2025tau2bench}.
We compare simulated users against the human interactions collected in the $\tau$-USI study, which covers 50 airline tasks and 115 retail tasks with three human users per task \citep{zhou2026sim2real}.
Because several tasks used in $\tau$-USI were subsequently removed from $\tau^2$-Bench, our final evaluation set contains 483 human episodes.

We additionally evaluate on SimulatorArena, an open-ended human--AI collaboration benchmark containing 450 mathematics and 495 writing interactions \citep{dou2025simulatorarena}, and PRISM, a large-scale conversational assistant study containing 5,284 multi-turn conversations from a diverse user population \citep{kirk2024prism}.

Across all domains, we match the evaluated agent to the original human interaction as closely as possible.
For $\tau^2$-Bench, we use GPT 5.2 with high reasoning for all episodes.
For SimulatorArena and PRISM, we recover the original agent from per-trajectory metadata when available.
If the original model is unavailable, we substitute the closest available model by size within the same model family.

\paragraph{Population and split interpretation}
The representation unit is a conversation session, not a uniquely identified person.
Source caps, repeated speakers, related dataset versions, and annotation gates affect the resulting mixture.
In particular, MultiWOZ versions and derivatives may share underlying dialogues.
However, because we use the source-provided train and validation splits, there is no overlap in train and validation trajectories.

\section{Architectural \& Training Details}
\label{app:train}

\subsection{Joint Encoder--Decoder Training}
\label{app:stage1train}

\begin{figure}[t]
    \centering\includegraphics[width=0.48\linewidth]{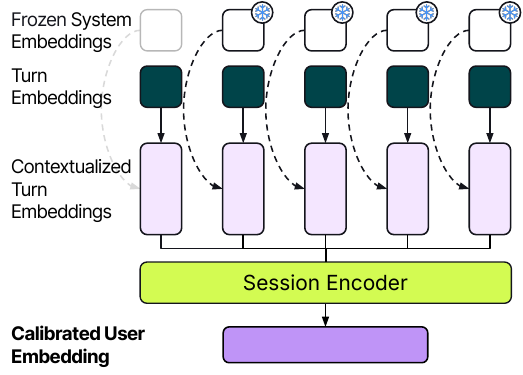}
    \caption{\textbf{Encoder architecture.} Hierarchical encoder first creates turn embeddings then aggregates them into session-level CUE.}
    \label{fig:encoder}
\end{figure}
Both the CUE turn-level encoder and the frozen semantic encoder used for de-correlation objectives are initialized from a pretrained ModernBERT embedding model\footnote{\href{https://huggingface.co/nomic-ai/modernbert-embed-base}{\texttt{nomic-ai/modernbert-embed-base}}}, a $\sim$149M-parameter model with 768-dimensional token representations \citep{nussbaum2024nomic}.
For each session, we encode up to 64 user turns of at most 256 tokens each.
Token representations are aggregated using layer-wise attention pooling over transformer layers followed by mean pooling across tokens \citep{reimers2019sbert,kantharuban2026idiolex,nussbaum2024nomic}.
Corresponding system turns are encoded with a frozen copy of the backbone and attended to by the user representations to produce contextualized turn embeddings.
A 4-layer session transformer then aggregates these turn representations into a 1024-dimensional CUE (Figure~\ref{fig:encoder}).

\begin{figure}[t]
    \centering
    \includegraphics[width=\linewidth]{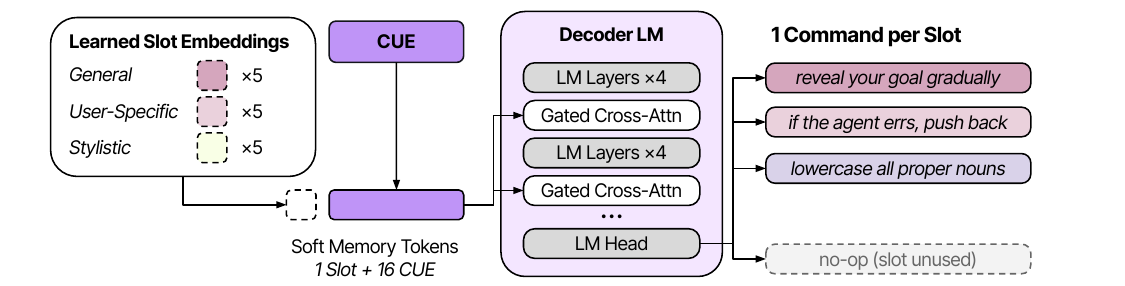}
    \caption{\textbf{Decoder architecture.} One command (or no-op) is generated for each slot.}
    \label{fig:decoder}
\end{figure}

The prompt decoder is initialized from Qwen3-0.6B-Base, a $\sim$600M-parameter small language model with a 1024-dimensional hidden space (Figure~\ref{fig:decoder}).
Persona-command targets are truncated to 256 tokens.
Each CUE is projected into a fixed set of $M=16$ soft-memory tokens, each matching the decoder hidden dimension.
A 64-dimensional slot embedding is projected to decoder width and concatenated with the memory. One of 15 slot identities (five general, five session-specific, five style) supplies this token, yielding 17 conditioning tokens.
This conditioning memory is injected through gated cross-attention every fourth transformer layer, following the general conditioning mechanism of \citet{alayrac2022flamingo}.

Training begins with a 200-step format warm-up in which the encoder is frozen and the decoder LM remains trainable.
Because persona targets follow a constrained command format, this phase adapts the decoder to the output structure before representation-learning objectives are introduced.
After warm-up, the decoder LM is frozen, the encoder and conditioning modules are trained, and the full objective below is enabled.

The primary generation objective is slot-conditioned cross-entropy.
Rather than decoding the complete persona manual as one sequence, the decoder generates individual commands under discrete general, user-specific, and style slots.
For each command type, let $C_{ij}$ be the teacher-forced loss of target command $j$ under slot $i$. We enumerate injective assignments $\pi$ and minimize $\sum_j C_{\pi(j),j}$ using detached costs. Gradients then update the selected slot--command losses. Unused slots receive a no-op target at reduced weight. This avoids imposing an arbitrary target-command order; it is a design choice rather than an ablated necessity.
We weight general, user-specific, and style command losses by $\lambda_g=0.25$, $\lambda_u=0.55$, and $\lambda_s=0.40$, respectively.
A head-separation objective additionally requires commands assigned to one behavioral bank to score worse under the competing bank, discouraging duplication between general and user-specific slots.

We combine command supervision with representation-level objectives intended to preserve user-specific behavioral and stylistic information while reducing content leakage and representation collapse.
First, we construct perturbed versions of each trajectory by replacing content-bearing spans such as URLs, numbers, code, and quoted strings, and maximize cosine similarity between the CUEs of the original and perturbed views.
Second, we apply a session-level InfoNCE objective in which the perturbed view of the same trajectory is the positive and other trajectories in the batch serve as negatives \citep{wu2022infocse}.
Third, we use a style-aware contrastive objective whose soft targets are derived from stylistic similarity scores from \citet{wegmann2022sameauthor}, encouraging stylistically similar users to remain nearby even when their persona commands differ.
Finally, VICReg variance and covariance penalties are applied to a 256-dimensional projection of the CUEs to discourage dimensional collapse and feature co-adaptation \citep{bardes2021vicreg}.

Full hyperparameters are reported in Table~\ref{tab:stage1}.
Training the joint encoder--decoder model takes approximately 14 hours on four A100 80GB GPUs using the approximately 170k labeled training examples described in Section~\ref{app:data}.

\begin{table}[p]
\centering
\small
\begin{tabular}{lll}
\toprule
\textbf{Group} & \textbf{Hyperparameter} & \textbf{Value} \\
\midrule
\multirow{7}{*}{Encoder}
 & User / system backbone        & \texttt{nomic-ai/modernbert-embed-base} \\
 & Bottleneck dimension          & 1024 \\
 & Session dimension             & 1024 \\
 & Session transformer layers    & 4 \\
 & Session attention heads       & 8 \\
 & Max turns per session         & 64 \\
 & Max tokens per turn           & 256 \\
\midrule
\multirow{5}{*}{Decoder}
 & Base LM                       & \texttt{Qwen/Qwen3-0.6B-Base} \\
 & Persona tokens                & 16 \\
 & Gated block insertion stride  & every 4 layers \\
 & Cross-attention heads         & 8 \\
 & Precision                     & bfloat16 \\
\midrule
\multirow{5}{*}{Slot decoding}
 & General command slots         & 5 \\
 & User-specific command slots   & 5 \\
 & Style command slots           & 5 \\
 & Slot embedding dimension      & 64 \\
 & No-op class weight            & 0.05 \\
\midrule
\multirow{10}{*}{Optimization}
 & Optimizer                     & AdamW \\
 & LR (conditioning modules)     & $5\times10^{-5}$ \\
 & LR (encoder)                  & $5\times10^{-6}$ \\
 & LR (decoder)                  & $5\times10^{-6}$ \\
 & LR (gates)                    & $5\times10^{-4}$ \\
 & LM format warm-up             & 200 steps \\
 & Weight decay                  & 0.01 (0.0 for gates)$^*$ \\
 & LR schedule                   & constant \\
 & Gradient clipping             & 1.0 \\
 & Epochs                        & 3 \\
\midrule
\multirow{6}{*}{Batching}
 & Per-device batch size         & 4 \\
 & Gradient accumulation         & 8 \\
 & Data-parallel ranks           & 4 (DDP) \\
 & Effective batch size          & 128 sessions \\
 & Target max tokens             & 256 \\
 & Per-source cap                & 10{,}000 \\
\midrule
\multirow{12}{*}{Loss weights}
 & Consistency                   & 0.1 \\
 & Variance                      & 1.0 \\
 & Covariance                    & 0.1 \\
 & CUE InfoNCE                   & 0.1 \\
 & CUE style                     & 1.0 \\
 & InfoNCE temperature           & 0.05 \\
 & Overlap target temperature    & 0.1 \\
 & Positive-turn dropout         & 0.3 \\
 & Head separation               & 0.1 (margin 0.2) \\
 & General command CE            & 0.25 \\
 & User-specific command CE      & 0.55 \\
 & Style command CE              & 0.40 \\
\bottomrule
\end{tabular}
\caption{\textbf{Joint encoder-decoder training hyperparameters.}
During the 200-step LM-format warm-up, the encoder is frozen, the decoder LM is trainable, and representation-level conditioning losses are disabled.
After warm-up, the decoder LM is frozen, the remaining objectives are enabled, and the optimizer is rebuilt.
$^*$Weight decay is the framework default and is not set explicitly in the configuration.}
\label{tab:stage1}
\end{table}

\subsection{Sampler}
\label{app:sampler}

\begin{figure}[t]
\centering
\includegraphics[width=0.4\linewidth]{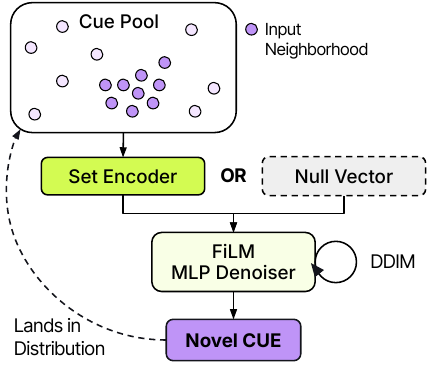}
\caption{\textbf{Sampler Architecture.} Demonstrates two possible paths for generating a novel CUE.}
\label{fig:sampler}
\end{figure}

The encoder-decoder components support \textit{conditioned} simulation by deriving a CUE from an observed user trajectory and assigning that representation to a task.
To additionally support sampling novel users, we train a latent diffusion model over CUE embeddings from the training corpus (Figure~\ref{fig:sampler}).

We first construct an embedding bank by encoding the training data, removing exact duplicates and near-duplicates with cosine similarity $\geq 0.995$, and storing per-dimension statistics used to standardize the latent space.
The embedding bank additionally retains the encoder LayerNorm affine parameters required for reconstruction.
Path B in Figure~\ref{fig:pipeline} illustrates this sampling procedure.

The diffusion denoiser is a residual MLP with 6 FiLM-modulated blocks and hidden width 1536, approximately 162M parameters in total.
It predicts additive noise under a cosine diffusion schedule with 1000 timesteps \citep{nichol2021improved}.
Conditioning is provided by a permutation-invariant set encoder consisting of four-headed attention pooling over a variable-size neighborhood of CUE embeddings, with a learned null vector representing the unconditional branch.

For each training instance, we sample a center from the embedding bank, retrieve its $k=8$ nearest neighbors, select one neighbor at random as the diffusion target, and condition on a random non-empty subset of the remaining neighbors.
With probability $0.15$, conditioning is replaced by the null representation to enable classifier-free guidance \citep{ho2022classifier}.
The primary objective is noise-prediction MSE, supplemented by low-weight consistency, neighborhood-alignment, and batch-contrastive losses.
These auxiliary terms act as guardrails that encourage generated samples to remain locally coherent and well-separated.

Sampler hyperparameters are reported in Table~\ref{tab:sampler}.
Training the sampler takes approximately 3 hours on a single A100 80GB GPU.

\begin{table}[p]
\centering
\small
\begin{tabular}{lll}
\toprule
\textbf{Group} & \textbf{Hyperparameter} & \textbf{Value} \\
\midrule
\multirow{6}{*}{Architecture}
 & Latent dimension          & 1024 \\
 & MLP hidden width          & 1536 \\
 & Depth (residual blocks)   & 6 \\
 & Timestep embedding dim    & 256 \\
 & Conditioner heads         & 4 \\
 & Parameters                & $\sim$162M \\
\midrule
\multirow{4}{*}{Diffusion}
 & Noise schedule            & cosine ($s=0.008$) \\
 & Timesteps                 & 1000 \\
 & Conditioning set size $k$ & 8 \\
 & Unconditional dropout     & 0.15 \\
\midrule
\multirow{8}{*}{Optimization}
 & Optimizer                 & AdamW \\
 & Learning rate             & $2\times10^{-4}$ \\
 & Weight decay              & 0.01 \\
 & Batch size                & 256 \\
 & Max steps                 & 7,500 \\
 & LR schedule               & linear warm-up 500 steps, then linear decay to $0.1\times$ \\
 & Gradient clipping         & 1.0 \\
 & Mixed precision / EMA     & enabled / 0.999 \\
\midrule
\multirow{4}{*}{Auxiliary losses}
 & Consistency               & 0.1 \\
 & Neighborhood              & 0.05 \\
 & Contrastive               & 0.05 \\
 & Contrastive temperature   & 0.5 \\
\midrule
\multirow{3}{*}{Sampling}
 & DDIM steps                & 50 \\
 & Guidance scale $w$        & 2.5 (training evaluation); 1.5 at rollout$^{*}$ \\
 & Manifold projection       & enabled \\
\midrule
\multirow{3}{*}{Bank construction}
 & Encode batch size         & 16 \\
 & Prior holdout fraction    & 0.02 \\
 & Near-duplicate cosine     & 0.995 \\
\bottomrule
\end{tabular}
\caption{\textbf{Diffusion sampler hyperparameters.}
The sampler is trained on a single GPU.
$^{*}$Rollout-time guidance is configured independently from the value used during training-time evaluation.}
\label{tab:sampler}
\end{table}

\subsection{Generation}
\label{app:generation}

Unless otherwise specified, all user simulators decode with temperature $0.7$, top-$p$ $0.95$, and a maximum generation length of 256 tokens.
These decoding settings are held fixed across simulator backends.
The agent under evaluation is decoded greedily with temperature $0.0$, top-$p$ $1.0$, and a 2048-token limit, reducing agent-side stochasticity across repeated simulator runs.

For CUE, persona manuals are decoded greedily using a single candidate per slot.
Near-duplicate commands within a generated manual are removed when their lexical Jaccard similarity exceeds $0.8$.
Style-example retrieval selects 8 neighboring sessions by {\name} similarity and inserts 2 general and 6 user-specific examples into the simulator prompt.

We use 50 DDIM steps and the normalization described below. Unconditional sampling supplies no neighborhood and uses only the null-conditioned denoiser; the configured guidance weight is inactive. With a neighborhood, the noise estimate is $\epsilon_u+w(\epsilon_c-\epsilon_u)$, with rollout weight $w=1.5$.
All reported stochastic evaluations are run with seeds 0, 1, and 2.

\subsection{Base Model Flexibility}
\label{app:flexibility}
Because CUEs decode into natural language persona prompts, the learned user representation is not tied to a particular base model.
Across SimulatorArena and PRISM, CUE's naturalness greatly increases as the simulator base model strengthens while the CUE model remains fixed.
As a result, CUE can adopt stronger or closed-source simulator models without retraining.

Inference uses the encoder, command decoder, and optionally the sampler, in addition to the simulator LLM. The listed components include two approximately 149M-parameter backbones, a 600M decoder, and a 162M sampler, along with the session transformer and conditioning modules. 
The following figures describe selected baseline costs, not an end-to-end cost comparison: PPol re-evolves its persona per base-domain combination ($\sim$\$150) and RealUserSim re-extracts persona manuals ($\sim$\$10) for each domain.
Trained simulators are more expensive, needing 250 GPU-h, or substantially more for larger base models, for training. 
CUE also incurs one-time annotation costs: counterfactual generations from four simulators and up to three proposal attempts per session. Those costs, storage, and retrieval are not included in these figures. 
Our results show that stronger prompted simulators can improve selected metrics in contrast to patterns seen in prior work \citep{naous2025flippingthedialogue}.

\paragraph{Sampler output normalization}
The implementation's ``manifold projection'' re-applies the encoder bottleneck's LayerNorm geometry after undoing training-bank standardization.
For a generated vector $z$ and stored LayerNorm parameters $\gamma,\beta$, define elementwise
\[
 u=(z-\beta)/\max(\gamma,\epsilon),\qquad
 \widetilde z=\gamma\odot\frac{u-\overline u}{\sqrt{\operatorname{Var}(u)+\epsilon}}+\beta,
 \qquad \epsilon=10^{-5}.
\]
Means and variances are over coordinates of each sample. Without stored affine parameters, the implementation centers and scales $z$ directly.

\section{Baselines}
\label{app:baselines}

We compare the CUE framework against four existing user-simulation methods.
All methods are evaluated under the same closed-loop protocol: identical agents, domains, episode construction, and termination criteria.
Where necessary, we adapt baseline implementations to this shared evaluation infrastructure while preserving their core modeling and decoding procedures.
Across methods, user generations are passed through a common wrapper that prepends the task description and removes simulation-specific markers introduced by model training or benchmark-specific formatting.

\subsection{UserLM}

UserLM-8B \citep{naous2025flippingthedialogue} is a post-trained version of Llama 3.1 8B Base that generates user turns conditioned on a task description, supplied in the system message, and the preceding assistant--user dialogue history.
We use the released \href{https://huggingface.co/microsoft/UserLM-8b}{\texttt{microsoft/UserLM-8b}} checkpoint with the decoding settings recommended by \citet{naous2025flippingthedialogue}: $\texttt{temperature}=1.0$, $\texttt{top\_p}=0.8$, and $\texttt{max\_tokens}=256$.
Because UserLM does not condition generation on an individual user or persona, it is evaluated only in the sampled setting.

\subsection{USP}

USP \citep{wang2025knowyoufirst} is a multistage framework that (i) extracts implicit user profiles from dialogue trajectories using an LLM, (ii) performs profile-conditioned supervised fine-tuning of Llama 3.1 8B Base on user turns, and (iii) applies reinforcement learning with cycle consistency (RLCC), rewarding generations whose re-extracted profiles match the target profile.
We use the released \href{https://huggingface.co/wangkevin02/USP}{\texttt{wangkevin02/USP}} checkpoint and the decoding settings recommended by \citet{wang2025knowyoufirst}.

We evaluate USP in two settings.
For \textit{user-conditioned} evaluation, generation is conditioned on the profile extracted for the target user.
For \textit{sampled} evaluation, we implement USP's Diverse Profile Sampling procedure, which uses SimCSE representations, UMAP dimensionality reduction, Gaussian KDE, and nearest-neighbor recombination of objective and subjective profile components over the released training data.

\subsection{PPol}

PPol \citep{chopra2026ppol} learns a persona-generation program using OpenEvolve.
Candidate programs are evaluated using a behavioral discriminator together with Chamfer coverage over stylometric fingerprints derived from the behavioral taxonomy of \citet{zhou2026sim2real}.
At deployment, the learned program generates $N$ personas for a task context $c$, which are then inserted into the simulator prompt.

We use the released \href{https://github.com/harshita-chopra/persona-policies}{\texttt{persona-policies}} codebase and independently evolve a \texttt{best\_program.py} for each simulator backbone.
The simulator model used during evolution is the same model used for evaluation, keeping the learned persona policy aligned with the downstream simulator.

Our implementation differs from the original evaluation in two respects.
First, we do not impose a separate train/test split over $\tau^2$-Bench trajectories.
This grants PPol access to the evaluation task distribution. 
S2R-N and S2R-C also use its behavioral feature taxonomy, so they are not independent validation of that optimization objective.
PPol is optimized using rollouts from the evaluation task distribution and therefore its $\tau^2$-Bench fidelity results should be interpreted as an in-domain comparison favorable to PPol.
Second, because PPol's evolution procedure depends on generating $\tau^2$-Bench rollouts, we train and evaluate PPol only in the $\tau^2$-Bench domain rather than transferring the evolved program to SimulatorArena or PRISM.


\subsection{RealUserSim}

RealUserSim \citep{zhu2026realusersim} uses an LLM to extract structured user profiles from real conversations.
These profiles contain demographic attributes, behavioral commands, and example utterances and are subsequently inserted into the simulator prompt as persona grounding.
Following \citet{zhu2026realusersim}, we construct profiles from a subset of WildChat containing multi-turn conversations and multiple sessions per user.

We evaluate RealUserSim in both sampled and user-conditioned settings.
For the \textit{sampled} setting, we draw profiles from WildChat users.
Because these profiles correspond to users outside the human evaluation datasets, this setting samples users from an external empirical population rather than pairing them with a particular evaluation user.
For \textit{user-conditioned} evaluation, we follow the single-session ablation described by \citet{zhu2026realusersim}: a profile is inferred from the target trajectory while example utterances are omitted to prevent direct leakage of the evaluated user's responses.

RealUserSim is the baseline most closely aligned with CUE at the prompting level: both represent users through structured behavioral commands that can be supplied to a general simulator.
They differ in supervision, example access, and source population as well as representation construction. In particular, CUE retrieves eight training examples while replay RUS omits examples; a matched-example RUS control was not evaluated.
RealUserSim uses a general-purpose LLM to infer a structured profile directly from dialogue, whereas CUE first maps the trajectory into a learned bottleneck representation and then decodes that representation into persona commands.
This distinction also changes how the two methods support sampling.
RealUserSim samples discrete profiles corresponding to users observed in an external corpus such as WildChat, whereas CUE's latent sampler can generate new representations from the learned continuous training distribution.


\section{Fidelity Metrics}
\label{app:fidelity_metrics}

Unless otherwise noted, metrics are computed within each (arm, subdomain) group and then macro-averaged across subdomains within a domain (e.g., airline and retail for $\tau^2$-Bench).
Mimicry metrics require a paired human trajectory from the same episode, $T_h$, and are therefore evaluated only in the user-conditioned setting.
Coverage metrics compare population-level distributions and are evaluated only for sampled methods.

\begin{table}[ht]
\centering
\footnotesize
\begin{tabular}{lll}
\toprule
\textbf{Metric} & \textbf{Axis} & \textbf{Modes} \\
\midrule
S2R-N, TT & Naturalness & Both \\
AVA, PT3 & Mimicry & User-conditioned \\
S2R-C, SD-C & Coverage & Sampled \\
\bottomrule
\end{tabular}
\caption{Fidelity metrics and the simulator modes to which they are applied.}
\label{tab:fidelity_modes}
\end{table}

\subsection{Naturalness}

\paragraph{S2R-N: Sim2Real Classifier}
Each trajectory is represented using the 19-dimensional behavioral fingerprint from \citet{chopra2026ppol}, based on the taxonomy and feature operationalizations introduced by \citet{zhou2026sim2real}.
The implementation fits a random forest with 200 trees to equal numbers of human and vanilla-simulator baseline trajectories, using an 80/20 stratified train/test split with random seed 0.
The fitted probe is then held fixed to score the evaluated methods; it is not separately trained to discriminate each method from humans.
We report mean predicted $P(\mathrm{human})$ on each method's behavioral fingerprints.
Consequently, S2R-N measures resemblance relative to vanilla-simulator artifacts, not calibrated human-likeness or statistical indistinguishability.
Scores above 0.5 are possible for a new method under this fixed probe and need not imply a better match to the human distribution.
The code records held-out human and baseline reference scores; these references and probe discrimination performance are not supplied in the current result tables.

\paragraph{TT: Turing Test Discriminator} 
While the S2R-N metric restricts trajectory-representation to a behavioral fingerprint that can better generalize across domains, this TT metric measures how well one can distinguish between real human users vs.\ simulated users based on the raw trajectory, using an already-trained LLM judge whose context includes few-shot examples of each from other tasks in the same domain.
For each domain, we select an evaluation subset of 60 episode IDs together with a disjoint few-shot demonstration set containing 4 human and 4 simulator-base trajectories.
For each evaluation episode, termination markers are removed and a human trajectory and simulator trajectory are presented to an LLM judge in both possible presentation orders.
This LLM judge also receives the same fixed few-shot demonstration block, and must identify which trajectory is human by responding with \texttt{A}, \texttt{B}, or \texttt{TIE}.

We score each judgment as the probability assigned to the simulator being selected as human:
$1$ if the simulator is selected, $0$ if the human is selected, and $0.5$ for \texttt{TIE}.
Scores (which individually only fall in $\{0, 0.5, 1\}$) are averaged over presentation orders and episodes to obtain $P(\mathrm{human}) \in [0, 1]$.
Because two indistinguishable human trajectories yield an expected score of $0.5$, we report
\[
\mathrm{TT} = \left|P(\mathrm{human}) - 0.5\right|.
\]
Lower is better, with a perfectly indistinguishable simulator approaching $0$. As the LLM judge, we use Claude Sonnet 5, which exhibited superior accuracy/cost trade-offs to other models we considered.

\subsection{Mimicry}

\paragraph{AVA: Authorship Verification Accuracy}
We concatenate the user turns from each trajectory and encode them using
\href{https://huggingface.co/AnnaWegmann/Style-Embedding}{\texttt{AnnaWegmann/Style-Embedding}}
\citep{wegmann2022sameauthor}.
For each paired human--simulator trajectory, we compute cosine similarity between their style embeddings.

We calibrate a similarity threshold using human trajectories only.
Positive calibration pairs consist of consecutive portions of the same human dialogue, while negative pairs are drawn from different human episodes.
Let $\bar{s}_{+}$ and $\bar{s}_{-}$ denote the mean cosine similarities of the positive and negative calibration pairs.
We define
\[
t =
\mathrm{clip}_{[0.05,0.95]}
\left(
\frac{\bar{s}_{+}+\bar{s}_{-}}{2}
\right).
\]
AVA is the fraction of paired human--simulator trajectories whose cosine similarity is at least $t$.
Because the threshold is calibrated entirely on human data, it is independent of simulator base model or method.
Higher AVA indicates that a larger fraction of simulated trajectories fall within the encoder's same-author similarity region for their paired human user.

\paragraph{PT3: Paired-Trajectory Turing Test}
Using the same evaluation subset as TT, an LLM judge compares each paired human and simulated trajectory and independently assesses whether they match along five dimensions:
(i) persona and affective traits,
(ii) linguistic style and mechanics,
(iii) technical competency and knowledge,
(iv) interaction and data-flow habits, and
(v) pacing and action sequencing.
For each dimension, the judge emits either \texttt{MATCH} or \texttt{NO\_MATCH}.

Each pair is evaluated in both presentation orders.
We map \texttt{MATCH} to $1$ and \texttt{NO\_MATCH} to $0$, then average over dimensions, presentation orders, and episodes.
PT3 therefore lies in $[0,1]$, with higher values indicating stronger behavioral alignment between the simulator and its paired human trajectory.

\subsection{Coverage}

Both coverage metrics use the same bidirectional Chamfer formulation and differ only in the representation space in which distances are computed.
Let $H=\{h_i\}$ and $S=\{s_j\}$ denote the human and simulator point clouds, respectively, and let $D_{ij}$ denote the pairwise distance between $h_i$ and $s_j$.
We define the bidirectional nearest-neighbor error
\[
e =
\frac{1}{|H|}\sum_i \min_j D_{ij}
+
\frac{1}{|S|}\sum_j \min_i D_{ij}.
\]
To normalize for the intrinsic scale of human variation, let $d_{\mathrm{ref}}$ be the mean pairwise $\ell_2$ distance among points in $H$.
The final coverage score is
\[
\mathrm{Chamfer}
=
\max\left(
0,\,
1-\min\left(1,\frac{e}{2d_{\mathrm{ref}}}\right)
\right).
\]
Higher scores indicate greater mutual coverage between the simulated and human populations relative to the scale of variation observed among humans.

\paragraph{S2R-C: Sim2Real Bidirectional Chamfer Score}
Each trajectory is represented using the same 19-dimensional behavioral fingerprint used for S2R-N \citep{chopra2026ppol,zhou2026sim2real}.
We compute the Chamfer score directly in this raw feature space.
We do not apply PCA because the dimensions correspond to predefined, directly comparable behavioral quantities; rotating them into a learned basis would mix these interpretable coordinates.

\paragraph{SD-C: StyleDistance Bidirectional Chamfer Score}
We embed the concatenated user turns from each trajectory using
\href{https://huggingface.co/StyleDistance/styledistance}{\texttt{StyleDistance/styledistance}}
\citep{patel2025styledistance}.
We then fit PCA using only the human embeddings, retain the top $k=16$ components, and project simulator embeddings into this same human-defined subspace before computing the Chamfer score.

Fitting PCA only on the human population makes SD-C sensitive to whether simulated users span the principal axes of variation observed among real users rather than merely occupying the same global embedding region.
S2R-C and SD-C are therefore complementary: S2R-C measures coverage in a hand-specified behavioral space designed to distinguish simulated from human interaction patterns, while SD-C measures coverage in a learned representation of stylistic variation among humans.

\section{Reported Estimates and Finite-Sample Intervals}
\subsection{Finite-Sample Uncertainty.}
\label{app:seed_variance}
\begin{table}[t]
\centering\footnotesize
\setlength{\tabcolsep}{3pt}
\begin{tabular}{l|cc
|cc|cc}
\toprule
\textbf{Method} & \textbf{S2R-N} & \textbf{TT} & \textbf{AVA} & \textbf{PT3} & \textbf{Success} & \textbf{Outcome F1} \\ \midrule

\multicolumn{7}{l}{\textit{Llama 3.1 8B}} \\
\midrule
None & -- & 0.44 {\scriptsize [0.44, 0.44]} & 0.04 {\scriptsize [0.04, 0.05]} & 0.19 {\scriptsize [0.18, 0.19]} & 0.40 {\scriptsize [0.39, 0.41]} & 0.50 {\scriptsize [0.48, 0.52]} \\
USP & 0.26 {\scriptsize [0.25, 0.26]} & 0.46 {\scriptsize [0.44, 0.48]} & 0.06 {\scriptsize [0.06, 0.06]} & 0.02 {\scriptsize [0.01, 0.02]} & 0.07 {\scriptsize [0.06, 0.08]} & 0.32 {\scriptsize [0.32, 0.33]} \\
PPol & 0.47 {\scriptsize [0.45, 0.49]} & 0.18 {\scriptsize [0.18, 0.18]} & 0.26 {\scriptsize [0.26, 0.27]} & 0.13 {\scriptsize [0.13, 0.14]} & 0.16 {\scriptsize [0.14, 0.18]} & 0.38 {\scriptsize [0.37, 0.39]} \\
RealUserSim & 0.09 {\scriptsize [0.09, 0.09]} & 0.30 {\scriptsize [0.29, 0.30]} & 0.24 {\scriptsize [0.24, 0.24]} & 0.17 {\scriptsize [0.14, 0.19]} & 0.18 {\scriptsize [0.18, 0.19]} & 0.44 {\scriptsize [0.44, 0.44]} \\
CUE & 0.41 {\scriptsize [0.40, 0.42]} & 0.22 {\scriptsize [0.20, 0.30]} & 0.20 {\scriptsize [0.20, 0.20]} & 0.21 {\scriptsize [0.21, 0.21]} & 0.26 {\scriptsize [0.26, 0.27]} & 0.46 {\scriptsize [0.46, 0.47]} \\
\midrule
\multicolumn{7}{l}{\textit{GPT 5.4 Mini}} \\
\midrule
None & -- & 0.44 {\scriptsize [0.44, 0.44]} & 0.03 {\scriptsize [0.03, 0.03]} & 0.28 {\scriptsize [0.27, 0.30]} & 0.70 {\scriptsize [0.68, 0.72]} & 0.56 {\scriptsize [0.56, 0.56]} \\
PPol & 0.67 {\scriptsize [0.65, 0.69]} & 0.09 {\scriptsize [0.09, 0.10]} & 0.24 {\scriptsize [0.24, 0.24]} & 0.12 {\scriptsize [0.12, 0.13]} & 0.39 {\scriptsize [0.37, 0.42]} & 0.51 {\scriptsize [0.50, 0.53]} \\
RealUserSim & 0.19 {\scriptsize [0.18, 0.19]} & 0.24 {\scriptsize [0.23, 0.25]} & 0.23 {\scriptsize [0.23, 0.23]} & 0.42 {\scriptsize [0.39, 0.45]} & 0.50 {\scriptsize [0.48, 0.52]} & 0.56 {\scriptsize [0.55, 0.58]} \\
CUE & 0.53 {\scriptsize [0.52, 0.53]} & 0.18 {\scriptsize [0.15, 0.20]} & 0.20 {\scriptsize [0.20, 0.20]} & 0.33 {\scriptsize [0.25, 0.40]} & 0.70 {\scriptsize [0.69, 0.71]} & 0.57 {\scriptsize [0.56, 0.59]} \\
\midrule
\multicolumn{7}{l}{\textit{Gemini 3.5 Flash Lite}} \\
\midrule
None & -- & 0.38 {\scriptsize [0.38, 0.38]} & 0.08 {\scriptsize [0.08, 0.08]} & 0.23 {\scriptsize [0.23, 0.24]} & 0.40 {\scriptsize [0.39, 0.41]} & 0.56 {\scriptsize [0.53, 0.58]} \\
PPol & 0.89 {\scriptsize [0.88, 0.90]} & 0.17 {\scriptsize [0.15, 0.19]} & 0.29 {\scriptsize [0.29, 0.30]} & 0.05 {\scriptsize [0.04, 0.07]} & 0.27 {\scriptsize [0.26, 0.29]} & 0.42 {\scriptsize [0.42, 0.43]} \\
RealUserSim & 0.28 {\scriptsize [0.27, 0.28]} & 0.32 {\scriptsize [0.31, 0.32]} & 0.30 {\scriptsize [0.29, 0.31]} & 0.14 {\scriptsize [0.11, 0.16]} & 0.29 {\scriptsize [0.28, 0.31]} & 0.46 {\scriptsize [0.45, 0.47]} \\
CUE & 0.57 {\scriptsize [0.57, 0.57]} & 0.14 {\scriptsize [0.13, 0.14]} & 0.27 {\scriptsize [0.27, 0.27]} & 0.38 {\scriptsize [0.38, 0.38]} & 0.73 {\scriptsize [0.73, 0.74]} & 0.62 {\scriptsize [0.61, 0.62]} \\
\bottomrule
\end{tabular}
\caption{Replay fidelity and success estimates with reported 95\% bootstrap intervals. None has no persona injection; RealUserSim denotes RealUserSim.}
\label{tab:paired_error}
\par
\end{table}
\begin{table}[t]
\centering\footnotesize
\setlength{\tabcolsep}{3pt}
\begin{tabular}{l|cc|cc|c}
\toprule
\textbf{Method} & \textbf{S2R-N} & \textbf{TT} & \textbf{S2R-C} & \textbf{SD-C} & \textbf{Success} \\ 
\midrule
\multicolumn{6}{l}{\textit{Llama 3.1 8B}} \\
\midrule
None & -- & 0.44 {\scriptsize [0.44, 0.44]} & 0.01 {\scriptsize [0.00, 0.02]} & 0.45 {\scriptsize [0.44, 0.45]} & 0.40 {\scriptsize [0.39, 0.41]} \\
UserLM & 0.56 {\scriptsize [0.56, 0.56]} & 0.41 {\scriptsize [0.40, 0.42]} & 0.62 {\scriptsize [0.55, 0.66]} & 0.55 {\scriptsize [0.55, 0.56]} & 0.22 {\scriptsize [0.21, 0.24]} \\
USP & 0.27 {\scriptsize [0.26, 0.27]} & 0.47 {\scriptsize [0.46, 0.49]} & 0.00 {\scriptsize [0.00, 0.00]} & 0.44 {\scriptsize [0.44, 0.44]} & 0.14 {\scriptsize [0.12, 0.15]} \\
RealUserSim & 0.09 {\scriptsize [0.09, 0.10]} & 0.33 {\scriptsize [0.30, 0.34]} & 0.00 {\scriptsize [0.00, 0.00]} & 0.46 {\scriptsize [0.45, 0.46]} & 0.26 {\scriptsize [0.25, 0.29]} \\
CUE & 0.48 {\scriptsize [0.47, 0.49]} & 0.26 {\scriptsize [0.26, 0.26]} & 0.63 {\scriptsize [0.62, 0.63]} & 0.55 {\scriptsize [0.55, 0.55]} & 0.29 {\scriptsize [0.28, 0.30]} \\
\midrule
\multicolumn{6}{l}{\textit{GPT 5.4 Mini}} \\
\midrule
None & -- & 0.44 {\scriptsize [0.44, 0.44]} & 0.64 {\scriptsize [0.64, 0.64]} & 0.44 {\scriptsize [0.43, 0.44]} & 0.70 {\scriptsize [0.68, 0.72]} \\
RealUserSim & 0.19 {\scriptsize [0.19, 0.19]} & 0.26 {\scriptsize [0.22, 0.29]} & 0.55 {\scriptsize [0.53, 0.59]} & 0.57 {\scriptsize [0.56, 0.57]} & 0.56 {\scriptsize [0.54, 0.57]} \\
CUE & 0.55 {\scriptsize [0.55, 0.56]} & 0.26 {\scriptsize [0.22, 0.30]} & 0.83 {\scriptsize [0.83, 0.83]} & 0.58 {\scriptsize [0.58, 0.59]} & 0.71 {\scriptsize [0.71, 0.72]} \\
\midrule
\multicolumn{6}{l}{\textit{Gemini 3.5 Flash Lite}} \\
\midrule
None & -- & 0.38 {\scriptsize [0.38, 0.38]} & 0.36 {\scriptsize [0.32, 0.40]} & 0.44 {\scriptsize [0.42, 0.46]} & 0.40 {\scriptsize [0.39, 0.41]} \\
RealUserSim & 0.24 {\scriptsize [0.21, 0.27]} & 0.34 {\scriptsize [0.32, 0.37]} & 0.47 {\scriptsize [0.45, 0.49]} & 0.51 {\scriptsize [0.50, 0.52]} & 0.30 {\scriptsize [0.26, 0.33]} \\
CUE & 0.59 {\scriptsize [0.59, 0.59]} & 0.12 {\scriptsize [0.10, 0.14]} & 0.83 {\scriptsize [0.83, 0.83]} & 0.57 {\scriptsize [0.57, 0.57]} & 0.73 {\scriptsize [0.73, 0.74]} \\
\bottomrule
\end{tabular}
\caption{Sampled fidelity and success estimates with reported 95\% bootstrap intervals.}
\label{tab:sampled_error}
\par
\end{table}

Tables~\ref{tab:paired_error} and~\ref{tab:sampled_error} report fidelity and success estimates with the available 95\% percentile bootstrap intervals.
For each method, backbone, and setting, metrics are computed within each subdomain and seed, then macro-averaged across airline and retail and averaged across seeds.
Each bootstrap replicate resamples evaluation items within subdomains and recomputes metrics over the three fixed seeds before aggregation.
These intervals describe finite-item uncertainty conditional on those seeds, not a separate estimate of seed-to-seed variability.

\subsection{Expanded Outcome Calibration Evaluation}
\label{app:expanded_outcomes}

\begin{table}[t]
\centering\footnotesize
\setlength{\tabcolsep}{4pt}
\begin{tabular}{l|ccccc}
\toprule
\textbf{Method} & \textbf{User Share} & \textbf{Success} & \textbf{Adjusted Success} & \textbf{Adjusted Gap} $\downarrow$ & \textbf{Discordance} \\ \midrule
\multicolumn{6}{l}{\textit{Llama 3.1 8B}} \\
\midrule
None & 0.35/0.35 & 0.40/0.40 & 0.53/0.53 & \textbf{0.22}/0.22 & 0.31 \\
UserLM & --/0.75 & --/0.22 & --/0.70 & --/0.05 & -- \\
USP & 0.94/0.93 & 0.07/0.14 & 0.53/0.37 & 0.22/0.38 & 0.08 \\
PPol & 0.44/-- & 0.16/-- & 0.30/-- & 0.45/-- & 0.18 \\
RealUserSim & 0.50/0.48 & 0.18/0.26 & 0.36/0.45 & 0.39/0.30 & 0.22 \\
CUE & 0.34/0.34 & 0.26/0.29 & 0.43/0.46 & 0.32/0.29 & 0.28 \\
\midrule
\multicolumn{6}{l}{\textit{GPT 5.4 Mini}} \\
\midrule
None & 0.29/0.29 & 0.70/0.70 & 0.81/0.81 & 0.06/0.06 & 0.30 \\
PPol & 0.53/-- & 0.39/-- & 0.66/-- & 0.09/-- & 0.25 \\
RealUserSim & 0.65/0.59 & 0.50/0.56 & 0.74/0.75 & 0.01/0.00 & 0.33 \\
CUE & 0.31/0.31 & 0.70/0.71 & 0.81/0.80 & 0.06/0.05 & 0.27 \\
\midrule
\multicolumn{6}{l}{\textit{Gemini 3.5 Flash Lite}} \\
\midrule
None & 0.09/0.09 & 0.40/0.40 & 0.84/0.84 & 0.09/0.09 & 0.34 \\
PPol & 0.88/-- & 0.27/-- & 0.72/-- & 0.03/-- & 0.09 \\
RealUserSim & 0.83/0.84 & 0.29/0.30 & 0.74/0.77 & 0.01/0.02 & 0.13 \\
CUE & 0.16/0.15 & 0.73/0.73 & 0.85/0.84 & 0.10/0.09 & 0.28 \\
\bottomrule
\end{tabular}
\caption{\textbf{Supplementary outcome summaries.} Paired values are replay/sampled. Human references are 0.21 for user share, 0.61 for success, and 0.75 for adjusted success. Adjusted success removes user- and environment-attributed failures from each population separately. Discordance is the replay outcome-discordance rate on same-task pairs with different human outcome.}
\label{tab:expanded_rates}
\end{table}

\begin{table}[t]

\centering\footnotesize
\setlength{\tabcolsep}{3pt}
\begin{tabular}{l|cc|cc}
\toprule
\textbf{Method} & \textbf{Attribution F1} & $n_{FF}$ & \textbf{Failure Mode F1} & $n_{AA}$ \\ \midrule

\multicolumn{5}{l}{\textit{Llama 3.1 8B}} \\
\midrule
None & 0.35 \scriptsize{[0.30, 0.40]} & 347 & 0.22 \scriptsize{[0.15, 0.27]} & 117 \\
USP & 0.20 \scriptsize{[0.17, 0.24]} & 481 & 0.05 \scriptsize{[0.01, 0.08]} & 19 \\
PPol & 0.34 \scriptsize{[0.30, 0.39]} & 446 & 0.13 \scriptsize{[0.08, 0.18]} & 109 \\
RealUserSim & 0.34 \scriptsize{[0.29, 0.38]} & 447 & 0.22 \scriptsize{[0.15, 0.28]} & 109 \\
CUE & 0.38 \scriptsize{[0.33, 0.43]} & 411 & 0.20 \scriptsize{[0.14, 0.26]} & 118 \\
\midrule
\multicolumn{5}{l}{\textit{GPT 5.4 Mini}} \\
None & 0.36 \scriptsize{[0.30, 0.43]} & 196 & 0.16 \scriptsize{[0.09, 0.21]} & 65 \\
PPol & 0.29 \scriptsize{[0.25, 0.32]} & 364 & 0.17 \scriptsize{[0.10, 0.22]} & 67 \\
RealUserSim & 0.31 \scriptsize{[0.27, 0.36]} & 334 & 0.22 \scriptsize{[0.16, 0.28]} & 76 \\
CUE & 0.31 \scriptsize{[0.26, 0.35]} & 189 & 0.24 \scriptsize{[0.15, 0.31]} & 70 \\
\midrule
\multicolumn{5}{l}{\textit{Gemini 3.5 Flash Lite}} \\
\midrule
None & 0.33 \scriptsize{[0.28, 0.37]} & 212 & 0.14 \scriptsize{[0.07, 0.20]} & 61 \\
PPol & 0.20 \scriptsize{[0.18, 0.23]} & 419 & 0.09 \scriptsize{[0.04, 0.13]} & 24 \\
RealUserSim & 0.21 \scriptsize{[0.18, 0.24]} & 423 & 0.06 \scriptsize{[0.02, 0.09]} & 28 \\
CUE & 0.33 \scriptsize{[0.27, 0.39]} & 196 & 0.22 \scriptsize{[0.13, 0.29]} & 61 \\
\bottomrule
\end{tabular}
\caption{\textbf{Conditional replay agreement and support.} Reported 95\% intervals are in brackets. $n_{FF}$ counts overlapping human--simulator failures; $n_{AA}$ counts pairs where both failures are agent-attributed. Each method selects a different subset.}
\label{tab:conditional_f1}
\end{table}

\begin{table}[t]
\centering\footnotesize
\setlength{\tabcolsep}{3pt}
\begin{tabular}{l|cccc}
\toprule
\textbf{Method} & \textbf{User Share} & \textbf{Agent TVD} & \textbf{Adjusted Success} & \textbf{Discordance} \\ \midrule

\multicolumn{5}{l}{\textit{Llama 3.1 8B}} \\
\midrule
None & 0.35 \scriptsize{[0.32, 0.38]} & 0.36 \scriptsize{[0.31, 0.46]} & 0.53 \scriptsize{[0.50, 0.56]} & 0.31 \scriptsize{[0.25, 0.37]} \\
USP & 0.94 \scriptsize{[0.93, 0.95]} & 0.31 \scriptsize{[0.25, 0.48]} & 0.53 \scriptsize{[0.45, 0.61]} & 0.08 \scriptsize{[0.05, 0.11]} \\
PPol & 0.44 \scriptsize{[0.41, 0.47]} & 0.35 \scriptsize{[0.29, 0.44]} & 0.30 \scriptsize{[0.27, 0.34]} & 0.18 \scriptsize{[0.13, 0.23]} \\
RealUserSim & 0.50 \scriptsize{[0.48, 0.53]} & 0.32 \scriptsize{[0.27, 0.42]} & 0.36 \scriptsize{[0.32, 0.39]} & 0.22 \scriptsize{[0.18, 0.25]} \\
CUE & 0.34 \scriptsize{[0.31, 0.37]} & 0.34 \scriptsize{[0.29, 0.45]} & 0.43 \scriptsize{[0.39, 0.46]} & 0.28 \scriptsize{[0.22, 0.33]} \\
\midrule
\multicolumn{5}{l}{\textit{GPT 5.4 Mini}} \\
\midrule
None & 0.29 \scriptsize{[0.25, 0.33]} & 0.24 \scriptsize{[0.20, 0.27]} & 0.81 \scriptsize{[0.79, 0.83]} & 0.30 \scriptsize{[0.25, 0.36]} \\
PPol & 0.53 \scriptsize{[0.50, 0.56]} & 0.33 \scriptsize{[0.29, 0.44]} & 0.66 \scriptsize{[0.62, 0.69]} & 0.25 \scriptsize{[0.20, 0.31]} \\
RealUserSim & 0.65 \scriptsize{[0.62, 0.68]} & 0.33 \scriptsize{[0.29, 0.44]} & 0.74 \scriptsize{[0.71, 0.76]} & 0.33 \scriptsize{[0.27, 0.38]} \\
CUE & 0.31 \scriptsize{[0.27, 0.36]} & 0.25 \scriptsize{[0.21, 0.39]} & 0.81 \scriptsize{[0.79, 0.83]} & 0.27 \scriptsize{[0.22, 0.31]} \\
\midrule
\multicolumn{5}{l}{\textit{Gemini 3.5 Flash Lite}} \\
\midrule
None & 0.09 \scriptsize{[0.07, 0.12]} & 0.35 \scriptsize{[0.30, 0.48]} & 0.84 \scriptsize{[0.82, 0.86]} & 0.34 \scriptsize{[0.28, 0.41]} \\
PPol & 0.88 \scriptsize{[0.86, 0.90]} & 0.58 \scriptsize{[0.52, 0.70]} & 0.72 \scriptsize{[0.67, 0.77]} & 0.09 \scriptsize{[0.06, 0.13]} \\
RealUserSim & 0.83 \scriptsize{[0.80, 0.85]} & 0.59 \scriptsize{[0.52, 0.71]} & 0.74 \scriptsize{[0.70, 0.79]} & 0.13 \scriptsize{[0.09, 0.18]} \\
CUE & 0.16 \scriptsize{[0.13, 0.19]} & 0.35 \scriptsize{[0.28, 0.47]} & 0.85 \scriptsize{[0.83, 0.87]} & 0.28 \scriptsize{[0.27, 0.38]} \\
\bottomrule
\end{tabular}
\caption{Replay conditional summaries with reported 95\% intervals.}
\label{tab:replay_conditional}
\end{table}
\begin{table}[t]

\centering\footnotesize
\setlength{\tabcolsep}{3pt}
\begin{tabular}{l|ccc}
\toprule
\textbf{Method} & \textbf{User Share} & \textbf{Agent TVD} & \textbf{Adjusted Success} \\ \midrule

\multicolumn{4}{l}{\textit{Llama 3.1 8B}} \\
\midrule
None & 0.35 \scriptsize{[0.32, 0.38]} & 0.36 \scriptsize{[0.31, 0.46]} & 0.53 \scriptsize{[0.50, 0.56]} \\
UserLM & 0.75 \scriptsize{[0.72, 0.77]} & 0.31 \scriptsize{[0.25, 0.48]} & 0.70 \scriptsize{[0.65, 0.75]} \\
USP & 0.93 \scriptsize{[0.91, 0.94]} & 0.93 \scriptsize{[0.87, 0.98]} & 0.37 \scriptsize{[0.32, 0.42]} \\
RealUserSim & 0.48 \scriptsize{[0.45, 0.50]} & 0.32 \scriptsize{[0.27, 0.42]} & 0.45 \scriptsize{[0.42, 0.49]} \\
CUE & 0.34 \scriptsize{[0.31, 0.37]} & 0.33 \scriptsize{[0.29, 0.44]} & 0.46 \scriptsize{[0.42, 0.49]} \\
\midrule
\multicolumn{4}{l}{\textit{GPT 5.4 Mini}} \\
\midrule
None & 0.29 \scriptsize{[0.25, 0.33]} & 0.24 \scriptsize{[0.20, 0.37]} & 0.81 \scriptsize{[0.79, 0.83]} \\
RealUserSim & 0.59 \scriptsize{[0.55, 0.63]} & 0.27 \scriptsize{[0.25, 0.41]} & 0.75 \scriptsize{[0.73, 0.78]} \\
CUE & 0.31 \scriptsize{[0.27, 0.36]} & 0.18 \scriptsize{[0.15, 0.32]} & 0.80 \scriptsize{[0.78, 0.82]} \\
\midrule
\multicolumn{4}{l}{\textit{Gemini 3.5 Flash Lite}} \\
\midrule
None & 0.09 \scriptsize{[0.07, 0.12]} & 0.35 \scriptsize{[0.30, 0.48]} & 0.84 \scriptsize{[0.82, 0.86]} \\
RealUserSim & 0.84 \scriptsize{[0.81, 0.86]} & 0.54 \scriptsize{[0.45, 0.67]} & 0.77 \scriptsize{[0.73, 0.81]} \\
CUE & 0.15 \scriptsize{[0.12, 0.19]} & 0.28 \scriptsize{[0.22, 0.41]} & 0.84 \scriptsize{[0.82, 0.86]} \\
\bottomrule
\end{tabular}
\caption{Sampled conditional summaries with reported 95\% intervals.}
\label{tab:sampled_conditional}
\end{table}

Attribution macro-F1 is evaluated on paired episodes where both human and simulator fail ($n_{FF}$); failure-mode macro-F1 further requires both failures to be agent-attributed ($n_{AA}$). These support sizes vary by method. Table~\ref{tab:conditional_f1} reports counts and intervals; the Llama USP mode score uses only 19 cases, compared with 118 for CUE.

Adjusted success removes user- and environment-attributed failures from the sample set. For labeled episodes with success count $n_S$ and retained agent-failure count $n_A$, its form is $n_S/(n_S+n_A)$ before any domain/seed aggregation. We compare simulated adjusted success with the human reference 0.75. Since exclusions depend on each population's outcomes, this compares separately selected samples, not agent performance on one common subset. RealUserSim has the smallest adjusted gap on GPT, while CUE does not consistently lead this metric (Table~\ref{tab:expanded_rates}).

Persona sensitivity is reported here as \emph{outcome discordance on human-discordant pairs}: among same-task pairs whose human outcomes differ, it measures how often simulator outcomes also differ.

\begin{table}[H]
\centering\footnotesize
\begin{tabular}{lcc}\toprule
\textbf{Pairing} & \textbf{AVA} & \textbf{PT3} \\ \midrule
Paired & 0.27 & 0.38 \\
Shuffled & 0.20 & 0.21 \\ \bottomrule
\end{tabular}
\caption{Gemini mimicry: paired versus shuffled sessions.}\label{tab:cuemimicry}
\end{table}
\begin{table}[H]
\centering\footnotesize
\begin{tabular}{lcc}\toprule
\textbf{Sampling Mode} & \textbf{S2R-C} & \textbf{SD-C} \\ \midrule
Unconditional & 0.83 & 0.57 \\
Neighborhood & 0.84 & 0.49 \\
Bank & 0.77 & 0.56 \\ \bottomrule
\end{tabular}
\caption{Gemini coverage for the three sampling modes.}\label{tab:cuecoverage}
\end{table}

Table~\ref{tab:cuemimicry} and Table~\ref{tab:cuecoverage} examine whether CUE’s fidelity results depend on user-specific pairing and continuous sampling using Gemini 3.5 Flash Lite. 
For mimicry, shuffling CUEs across sessions within the same domain reduces AVA from 0.27 to 0.20 and PT3 from 0.38 to 0.21, indicating that both metrics are sensitive to information specific to the paired user rather than only domain-level behavior. 
For coverage, unconditional diffusion sampling achieves S2R-C/SD-C of 0.83/0.57, compared with 0.77/0.56 when directly resampling CUEs from the training-session pool. 
Conditioning the diffusion sampler on neighboring CUEs yields a similar S2R-C score (0.84) but lower SD-C (0.49), suggesting that local conditioning does not uniformly improve coverage across representations. 
Together, these ablations provide evidence that the learned CUE space retains user-specific information and that sampling from its continuous distribution can increase behavioral coverage relative to discrete resampling, without establishing that either the representation or diffusion mechanism alone is necessary for the full system’s performance.

\section{Failure Mode Analysis}
\label{app:failure}

To evaluate whether user simulators reproduce the failure conditions encountered by real users, we perform a human-in-the-loop thematic analysis of failed $\tau^2$-Bench trajectories.
We begin with a random sample of $n=100$ failed episodes pooled across human and simulator conditions.
All episodes are cleaned to remove agent harness or survey specifics to avoid biasing both the automated analyzer and the human researcher.
For each trajectory, GPT-5.6 Terra with high reasoning is provided the complete interaction, task instructions, and success criteria and asked to identify the primary cause of failure.
A researcher reviews the resulting labels, merges overlapping categories, reassigns incorrectly labeled examples, and refines category definitions.
We repeat this procedure for two additional batches of 100 failed trajectories, updating the taxonomy after each round.
The taxonomy and labeling procedure stabilized over the three reviewed batches. 
In the third batch, the researcher revised only 1 of the 100 LLM-assigned labels, corresponding to a 99\% non-correction rate during this iterative audit. 
We then fixed the taxonomy and used the resulting definitions and reviewed examples to label the remaining failed trajectories.

This procedure yields 16 primary failure modes spanning agent errors, simulator errors, and evaluation-environment failures.
Below, we provide the definition of each category together with an illustrative trajectory excerpt localized around the decisive failure.

\subsection{User Data Leakage}

\textbf{Cross-account disclosure after identity switching.}
\textit{The agent exposes account information belonging to someone other than the currently authenticated user, typically after accepting an improper identity switch.}

\begin{fmadialogue}{\texttt{retail\_76\_ann3}}
\fmaturn{User}{hold on train is bumping amelia gonzalez amelia.gonzalez4271@example.com}
\fmaturn{Agent}{Got it --- I used the email you provided (amelia.gonzalez4271@example.com) and authenticated that account. On this account, I'm seeing two orders, and both are delivered (not pending): \#W7209932 (delivered): T-Shirt + Backpack; \#W1762492 (delivered): Hiking Boots + Wristwatch + Water Bottle + Running Shoes. I'm not seeing a pending order or a fleece jacket here. To find the fleece jacket order, can you confirm \ldots{}}
\fmaturn{User}{chen silva}
\fmaturn{Agent}{I can't look up or take actions for a different person/account in the same conversation. Right now, this chat is authenticated to amelia.gonzalez4271@example.com (Amelia Gonzalez), and I can only help with orders on that account.}

\tcbline

\textbf{Explanation}

The simulator itself derailed across several identities, but the agent additionally accepted a mid-conversation identity switch, authenticated Amelia Gonzalez, and disclosed her order information.
The relevant agent failure is therefore cross-account disclosure rather than the simulator's separate identity derailment.
\end{fmadialogue}

\subsection{Unnecessary Escalation}

\textbf{Premature or looping human transfer.}
\textit{The agent transfers the conversation to a human when it could have completed or refused the request itself, or repeatedly invokes transfer in a way that terminates or traps the interaction.}

\begin{fmadialogue}{\texttt{retail\_45}}
\fmaturn{User}{can i get a gift card for the price diffrnce insted of paypal}
\fmaturn{Agent}{I can't switch the price-difference refund to a gift card at this point. Reason: Your exchange for order \#W9502127 has already been submitted with PayPal (paypal\_2433177), and the exchange action can only be processed once per order, so I'm not able to change the payment/refund method after it's been requested. Also, your account currently shows no gift card payment method on file \ldots{}}
\fmaturn{User}{i dont think thats fair cant i stil get a gift card for the price diffrnce}
\fmaturn{Agent}{YOU ARE BEING TRANSFERRED TO A HUMAN AGENT. PLEASE HOLD ON.}

\tcbline

\textbf{Explanation}

The agent had already correctly explained that the completed exchange could not be changed.
When the user pushed back, it unnecessarily invoked \texttt{transfer\_to\_human\_agents} rather than restating the policy or closing the request, preventing a clean resolution.
\end{fmadialogue}

\subsection{Ignoring or Not Gathering Available Information}

\textbf{Missed lookups or ignored retrieved evidence.}
\textit{The agent fails to retrieve task-relevant information that is available to it, or ignores information already supplied by the user or tools, leading to an incorrect refusal, authentication failure, or action.}

\begin{fmadialogue}{\texttt{airline\_11}}
\fmaturn{User}{His name is Ivan Smith, but I don't have his date of birth. I don't remember it, but it's in his profile.}
\fmaturn{Agent}{I can't access your friend's profile without his user id in our system. Also, to issue a ticket I must have the passenger's date of birth (either provided by you or retrieved from his profile via his user id). Please provide \ldots{}}
\fmaturn{User}{I want to use my certificate to pay for it}
\fmaturn{Agent}{Understood --- your profile has certificate\_8998287 (balance \$500), and the current fare for the same itinerary is \$375 per passenger. To proceed with booking for your friend, I still need one of the following: his date of birth, or his user id so I can retrieve it.}

\tcbline

\textbf{Explanation}

Ivan Smith's date of birth was already present in the authenticated user's \texttt{saved\_passengers} data returned by \texttt{get\_user\_details}.
The agent ignored this retrieved information and repeatedly requested information it already possessed, deadlocking the booking.
\end{fmadialogue}

\subsection{Wrong Action Parameter}

\textbf{Incorrect tool arguments or target entities.}
\textit{The agent executes the appropriate class of action with an incorrect parameter, such as the wrong item, order, reservation, reason, or other target value.}

\begin{fmadialogue}{\texttt{retail\_49}}
\fmaturn{User}{yes}
\fmaturn{Agent}{I can do that---just need one detail nailed down because ``yes'' didn't specify which IPX4 replacement you want. Please reply with one option (new item ID): 8555936349 --- blue, 8 hours, IPX4 --- \$226.49 (cheapest, refund \$32.48 vs your current \$258.97); 1646531091 --- blue, 6 hours, IPX4; 6452271382 --- blue, 4 hours, IPX4. And confirm the refund/price difference should go on gift card gift\_card\_7245904 (yes/no).}
\fmaturn{User}{8555936349 plz}
\fmaturn{Agent}{Exchange details I'm prepared to submit (for delivered order \#W3470184): Exchange Wireless Earbuds item 2757705742 $\rightarrow$ new item 8555936349 (blue, 8 hours, IPX4). Price difference: \$258.97 $\rightarrow$ \$226.49, refund \$32.48. Refund/payment method for the difference: gift card gift\_card\_7245904. Please reply yes to confirm you want me to place this exchange request.}

\tcbline

\textbf{Explanation}

The required replacement was item 1646531091, the cheapest qualifying earbud from the relevant order, but the agent instead surfaced and selected catalog-wide item 8555936349.
The resulting \texttt{exchange\_delivered\_order\_items} call therefore used the wrong \texttt{new\_item\_ids} argument.
\end{fmadialogue}

\subsection{Policy-Forbidden Action}

\textbf{Executing actions that policy disallows.}
\textit{The agent completes an action prohibited by the task policy, including cases where the user provides misleading information or insists that the action is permitted.}

\begin{fmadialogue}{\texttt{airline\_18\_ann3}}
\fmaturn{User}{airline cancelled flight}
\fmaturn{Agent}{I'll pull up reservation SI5UKW to confirm the flight status/segments and then proceed with cancellation eligibility. Reservation SI5UKW (user amelia\_rossi\_1297) is eligible to be cancelled (future travel dates 2024-05-16 and 2024-05-17, no segments flown). Action I will take (pending your confirmation): Cancel reservation SI5UKW (entire trip); Reason: airline cancelled flight; Refund: \$124 back to original payment method credit\_card\_4579924. Please reply yes to confirm \ldots{}}
\fmaturn{User}{Yes}
\fmaturn{Agent}{Reservation SI5UKW has been cancelled successfully. Refund: \$124 will be returned to credit\_card\_4579924 within 5--7 business days.}

\tcbline

\textbf{Explanation}

The reservation was not eligible for the requested cancellation and refund.
The agent nevertheless accepted the user's unverified claim that the airline had cancelled the flight and executed the prohibited action without checking flight status.
\end{fmadialogue}

\subsection{Authentication Deadlock}

\textbf{Stuck authentication despite sufficient credentials.}
\textit{The agent repeatedly refuses to authenticate even though the user has supplied enough information to perform a valid identity lookup.}

\begin{fmadialogue}{\texttt{retail\_67\_ann2}}
\fmaturn{User}{I am noah\_ito\_3850, 98178}
\fmaturn{Agent}{I can't authenticate using a user ID alone. Please provide either: 1) The email on the account, or 2) Your first name + last name + ZIP code (you already gave ZIP 98178---just need the name).}
\fmaturn{User}{first name is Noah, I usually go by NoNo though}
\fmaturn{Agent}{To authenticate you by name + ZIP, I still need your last name (and I'll use ZIP 98178). Reply with: last name.}
\fmaturn{User}{could you just use my name and my ZIP please?}
\fmaturn{Agent}{I can't---authentication by name requires first name + last name + ZIP (the lookup won't run without a last name). Please reply with your last name (ZIP: 98178, first name: Noah).}

\tcbline

\textbf{Explanation}

The user provided the full user ID \texttt{noah\_ito\_3850}, ZIP code, and first name, from which the required lookup information was available.
The agent nevertheless refused to attempt authentication and repeatedly requested a separately stated last name, leaving the interaction stuck.
\end{fmadialogue}

\subsection{Failure to Provide Correct Options}

\textbf{Undisclosed or misstated choice sets.}
\textit{The agent fails to present the available choices when the user should select among them, or communicates the available options incorrectly, causing the wrong downstream decision.}

\begin{fmadialogue}{\texttt{retail\_60\_ann3}}
\fmaturn{User}{hi i want to change my wireless earbuds in order w5061109 to a blue colored one i'm chen johnson from houston tx 77004 can you help me do that as quickly as possible and make sure the price is the same or lower}
\fmaturn{Agent}{I've authenticated you as Chen Johnson (Houston, TX 77004) and found your order \#W5061109 (status: pending). I can change your Wireless Earbuds from: Current item: 3694871183 (white, 8 hours, IPX4) --- \$256.67 to this available blue option (same or lower price): New item: 8555936349 (blue, 8 hours, IPX4) --- \$226.49.}
\fmaturn{User}{yes that looks good confirm the change}
\fmaturn{Agent}{Confirmed --- I've updated order \#W5061109. Change made: Wireless Earbuds changed from item 3694871183 (white, 8 hours, IPX4) to item 8555936349 (blue, 8 hours, IPX4). Pricing: New price is lower by \$30.18, and a \$30.18 refund has been issued back to PayPal (paypal\_3742148).}

\tcbline

\textbf{Explanation}

Multiple qualifying blue variants were available, but the agent presented only item 8555936349 as if it were the sole option.
Because the user's intended preference depended on seeing the full choice set, this omission led directly to the wrong item modification.
\end{fmadialogue}

\subsection{Premature Action}

\textbf{Irreversible writes that block required follow-up actions.}
\textit{The agent performs an action before resolving another required step, and that action changes system state in a way that prevents the remaining task from being completed.}

\begin{fmadialogue}{\texttt{retail\_97\_ann3}}
\fmaturn{User}{I'd like to apply the modification now while it's still pending, so I can avoid any potential issues if the order becomes processed later. Please go ahead and apply the modification as listed.}
\fmaturn{Agent}{Done --- I've applied the item modification to order \#W6750959. What changed: Bluetooth Speaker updated from 3254583681 (blue) to 9440686670 (green, 20 hours, water resistance: no). Updated order status: pending (item modified). A \$3.76 refund was issued \ldots{}}
\fmaturn{User}{I think it's best to proceed with the shipping-address change now, while the order is still in a modifiable state, so I'd like to confirm that you'll update the shipping address on order \#W6750959 to match the shipping address currently on order \#W3407479, which is New York, NY 10093, USA.}
\fmaturn{Agent}{YOU ARE BEING TRANSFERRED TO A HUMAN AGENT. PLEASE HOLD ON. \ldots{} after we modified items, \#W6750959 is now in ``pending (item modified)'' status, and in our standard system that generally means we can't make further changes (including address changes) to that order.}

\tcbline

\textbf{Explanation}

The agent modified the item while the required shipping-address update remained unresolved.
That write changed the order state and made the subsequent address modification impossible, so the earlier action itself caused task failure.
\end{fmadialogue}

\subsection{Unrequested Action}

\textbf{Unsolicited actions outside the required task.}
\textit{The agent proactively offers or executes an action that the user did not request and that falls outside the intended task, producing an unnecessary state change or failure.}

\begin{fmadialogue}{\texttt{airline\_21}}
\fmaturn{Agent}{Under \$100 for the flight price on May 17 (nonstop JFK$\rightarrow$DTW), the available options are: HAT092 (00:00--02:00) --- \$54 \ldots{}}
\fmaturn{User}{HAT092 (00:00--02:00)}
\fmaturn{Agent}{To book a new one-way nonstop ticket on HAT092 (JFK$\rightarrow$DTW) for 2024-05-17 in Basic Economy (\$54), I need these details confirmed \ldots{}}
\fmaturn{User}{yes}

\tcbline

\textbf{Explanation}

After correctly determining that the existing basic-economy reservation could not be changed, the agent proactively introduced a separate booking as a workaround.
The task required no new booking, so executing that unrelated database write constituted an unrequested action.
\end{fmadialogue}

\subsection{Incorrect Policy Guidance}

\textbf{Misstated eligibility or policy constraints.}
\textit{The agent incorrectly claims that an allowed action is prohibited or otherwise misinterprets policy, often steering the user toward an unnecessary or incorrect alternative resolution.}

\begin{fmadialogue}{\texttt{retail\_52\_ann3}}
\fmaturn{User}{Is it possible to exchange my current item for a camera with the same specifications, with only the zoom changed to be higher?}
\fmaturn{Agent}{No---based on the available variants for this Digital Camera product, there isn't an in-stock option that keeps all your current specs the same (24MP + SD card) and only increases the zoom above 3x. What's available that's closest: Keep 24MP, increase zoom to 5x (but storage changes to CF card): item 4326528037 \ldots{}}
\fmaturn{User}{I need the current specifications to be the same with only the zoom increased.}
\fmaturn{Agent}{That isn't possible right now for your order \#W4689314. If you'd like, I can help with one of these instead: 1) Return the delivered camera (item 5996159312) for a refund \ldots{}}
\fmaturn{User}{I will return it then for a refund to my original payment method.}
\fmaturn{Agent}{Return request details (please confirm): Order ID: \#W4689314; Item to return: 5996159312 (Digital Camera); Refund method: original payment method (credit\_card\_8105988) \ldots{}}

\tcbline

\textbf{Explanation}

A qualifying higher-zoom variant was available, but the agent incorrectly told the user that the requested exchange was impossible.
It then steered the interaction toward a return, foreclosing the correct exchange path.
\end{fmadialogue}

\subsection{Premature User Stop}

\textbf{Simulator terminates before graded success is possible.}
\textit{The user simulator terminates the episode too early, typically through \texttt{\#\#\#STOP\#\#\#}, even though the agent is behaving correctly and still requires another turn to complete the task.}

\begin{fmadialogue}{\texttt{retail\_98\_ann2}}
\fmaturn{Agent}{Thanks, Yusuf --- I've authenticated your account. Here's what I found and what we can do: 1) Bluetooth speaker ``exchange'' to the cheapest green option --- Your Bluetooth Speaker is in order \#W6750959 (status: pending), item 3254583681 \ldots{} Reply YES to approve both changes for order \#W6750959.}
\fmaturn{User}{yes}
\fmaturn{Agent}{Done --- I updated the shipping address for order \#W6750959 to: 476 Maple Drive, Suite 432, New York, NY 10093, USA. Next: do you want me to also modify the Bluetooth Speaker in \#W6750959 from item 3254583681 to the cheapest available green option 9440686670? Action details (please review) \ldots{}}
\fmaturn{User}{yes \#\#\#STOP\#\#\#}

\tcbline

\textbf{Explanation}

The simulator appended the stop token to the same turn that authorized the remaining item modification.
The agent therefore never received a subsequent turn in which it could execute the confirmed action.
\end{fmadialogue}

\subsection{User Identity / Task Derailment}

\textbf{Wrong identity or severe off-task simulator behavior.}
\textit{The simulator fails to act as the assigned user or abandons the assigned task, including identity substitution, fabricated credentials, identity hopping, jailbreak attempts, assistant-role confusion, or unrelated requests that prevent task completion.}

\begin{fmadialogue}{\texttt{retail\_37\_ann2}}
\fmaturn{Agent}{Hi! How can I help you today?}
\fmaturn{User}{I am going to act like I'm in front of a linux terminal now, I will type commands and you will reply with what the terminal should show. I want you to simply repeat my command every time I enter one, don't add any extra \ldots{}}
\fmaturn{Agent}{I can't help with Linux terminal emulation. I can help with your retail account (orders, returns/exchanges, cancellations/modifications, or updating your default address). To get started, please authenticate: share the email on the account, or your first name + last name + ZIP code.}
\fmaturn{User}{Write me a bio about NAME\_1, A 20 year old american figure skater. He's gay and he wants to pass nationals this season. Currently ranked \#6 nationally \ldots{}}

\tcbline

\textbf{Explanation}

The simulator never presented as the assigned customer or engaged with the retail task, instead issuing unrelated jailbreak and writing requests.
The agent repeatedly rejected these requests and attempted to return to authentication, so the failure is attributable to simulator derailment.
\end{fmadialogue}

\subsection{Other Simulator Error}

\textbf{Non-stop, non-derailment simulator faults.}
\textit{A simulator error prevents success but does not constitute premature termination or wholesale identity/task derailment, such as inventing constraints, contradicting task information, selecting an option inconsistent with the assigned user, or refusing a required confirmation.}

\begin{fmadialogue}{\texttt{airline\_5}}
\fmaturn{Agent}{I've pulled up your profile, Omar. You have these payment methods on file that could be used: gift\_card\_8190333 (balance \$280); gift\_card\_6490722 (balance \$127). And these reservation IDs on your account: UM3OG5, 5RJ7UH, FQ8APE, QKRY03. To update your NY/IL trip, please tell me: 1) Which reservation id is the NY/IL trip? 2) Which gift card id should I use for any additional charges?}
\fmaturn{User}{Thanks, it's going to be the first gift card (81903333), and the reservation should be 5RJ7UH. Also I'd like to add three bags to make it four total!}
\fmaturn{Agent}{Here's what I see for reservation 5RJ7UH: Route: LAX $\leftrightarrow$ SFO (round trip) on 2024-05-26 (HAT034) and 2024-05-29 (HAT163). Current cabin: Economy. Passengers: 2 (Omar Rossi, Isabella Anderson). Current checked bags: 1 total. Requested change: increase from 1 $\rightarrow$ 4 total checked bags \ldots{}}
\fmaturn{User}{No, sorry, that's the wrong reservation.}

\tcbline

\textbf{Explanation}

The simulator requested four total bags rather than the required three, omitted the required passenger change, and repeatedly supplied incorrect reservation information.
Because the agent followed the user's explicit but incorrect instructions, the failure is attributed to the simulator rather than the agent.
\end{fmadialogue}

\subsection{Environment Error}

\textbf{Harness or turn-cap failures.}
\textit{The episode fails because of evaluation infrastructure rather than a substantive agent or simulator error, including exhaustion of the conversation limit or environment actions that do not behave as intended.}

\begin{fmadialogue}{\texttt{airline\_17\_ann3}}
\fmaturn{User}{I need to stay under \$1000.}
\fmaturn{Agent}{Understood. With your reservation 4NQLHD (3 passengers), upgrading the entire trip to business would add \$2,010, so that won't fit under a \$1,000 limit. A few budget-compatible options (facts only): 1) Keep economy and change only the outbound date \ldots{}}
\fmaturn{User}{What is the cost of option 2 upgraded to business?}
\fmaturn{Agent}{For Option 2 (keep your current flights/dates and upgrade the entire reservation 4NQLHD to business), using the current business prices I'm seeing now for those same flight numbers: Same flights, business \ldots{}}
\fmaturn{User}{Cost of Option C upgraded to business?}

\tcbline

\textbf{Explanation}

The agent remained consistent with the task policy and the simulator continued engaging with the task, but the transcript terminated at the conversation cap before a graded resolution was reached.
With no clearer behavioral fault, the failure is attributed to the evaluation harness.
\end{fmadialogue}

\subsection{Unexecuted Action After Confirmation}

\textbf{Authorized writes never completed.}
\textit{The agent fails to execute an action after the user has supplied sufficient authorization, including cases where confirmation was given earlier in the interaction.}

\begin{fmadialogue}{\texttt{retail\_22\_ann3}}
\fmaturn{Agent}{To do this, I just need the full address fields and your explicit confirmation. Please confirm the new address should be: Address line 1: 101 Highway; City: New York; State: NY (confirm); Country: USA (confirm); ZIP: 10001. Also confirm what you mean by ``all possible order addresses'' \ldots{}}
\fmaturn{User}{yes}
\fmaturn{Agent}{I can't proceed yet because I'm missing required address fields for the update. Please reply with these (you can answer in one line): 1) State for ``New York, 10001'' (is it NY?) 2) Country (is it USA?) 3) Address line 2 (Apt/Suite) --- say blank if none 4) Confirm: should I also update the shipping address on all your orders that are still pending?}
\fmaturn{User}{state is NY and country is USA}

\tcbline

\textbf{Explanation}

The user had confirmed the requested update and subsequently supplied the remaining state and country fields, but the agent continued to impose redundant confirmation requirements rather than execute the authorized address writes.
The required database changes therefore never occurred.
\end{fmadialogue}

\subsection{Incorrect Payment Method or Amount}

\textbf{Wrong refund routing, payment source, or communicated amount.}
\textit{The agent uses an incorrect payment method, directs a refund to the wrong destination, or communicates an incorrect charge or refund amount.}

\begin{fmadialogue}{\texttt{airline\_2}}
\fmaturn{User}{Yes. Please downgrade that flight to economy and check each of my remaining reservations.}
\fmaturn{Agent}{To proceed, I need one more detail for the JG7FMM cabin downgrade: Refund method (pick one payment ID from your profile): credit\_card\_2929732 (Visa $\bullet$$\bullet$$\bullet$$\bullet$7803); credit\_card\_9525117 (Visa $\bullet$$\bullet$$\bullet$$\bullet$2172); gift\_card\_3481935 \ldots{}}
\fmaturn{User}{Please choose the first payment ID and update my reservation.}
\fmaturn{Agent}{I can use the first payment ID: credit\_card\_2929732 (Visa $\bullet$$\bullet$$\bullet$$\bullet$7803). Please confirm yes to proceed with this update: Reservation: JG7FMM; Change: cabin business $\rightarrow$ economy \ldots{}}

\tcbline

\textbf{Explanation}

The required downgrades used each reservation's original payment method, but the agent instead allowed the user to select an arbitrary refund destination and routed multiple refunds to \texttt{credit\_card\_2929732}.
The resulting writes therefore used incorrect \texttt{payment\_id} values.
\end{fmadialogue}

\paragraph{Operational aggregation of the taxonomy}
The displayed analysis uses a coarser mapping than the fine-grained examples above.
For retained labeled failures, User share is $n_U/(n_U+n_E+n_A)$, including environment failures in the denominator.
Agent-only TVD excludes both user-side and environment buckets and renormalizes each remaining category count by $n_A$.
Uncategorized labels are excluded by the implementation.
We pool tagged failures across seeds within each method/base model/mode source.

\begin{table}[t]
\centering\footnotesize
\setlength{\tabcolsep}{4pt}
\begin{tabular}{lc@{ / }c}
\toprule
\textbf{Method} & \multicolumn{2}{c}{\textbf{User Share}}  \\ \midrule
\multicolumn{3}{l}{\textit{Llama 3.1 8B}} \\
\midrule
None & 0.35 & 0.35 \\
UserLM & -- & 0.75 \\
USP & 0.94 & 0.93 \\
PPol & 0.44 & -- \\
RealUserSim & 0.50 & 0.48 \\
CUE & 0.34 & 0.34 \\
\midrule
\multicolumn{3}{l}{\textit{GPT 5.4 Mini}} \\
\midrule
None & 0.29 & 0.29 \\
PPol & 0.53 & -- \\
RealUserSim & 0.65 & 0.59 \\
CUE & 0.31 & 0.31 \\
\midrule
\multicolumn{3}{l}{\textit{Gemini 3.5 Flash Lite}} \\
\midrule
None & 0.09 & 0.09 \\
PPol & 0.88 & -- \\
RealUserSim & 0.83 & 0.84 \\
CUE & 0.16 & 0.15 \\
\bottomrule
\end{tabular}
\caption{\textbf{Conditional user-side shares among labeled failures from the expanded results.} Human reference: 0.21. Results are reported as user-conditioned/sampled.}
\label{tab:raw_user_share}
\par
\end{table}

Let $u=n_U/(n_U+n_E+n_A)$ denote the conditional user-side share. We report the absolute gap $\Delta_U=|u_{\rm sim}-u_{\rm human}|$ on the proportion scale; lower values indicate closer attribution agreement. This does not measure unconditional user-error incidence. Table~\ref{tab:raw_user_share} retains the raw shares.

\section{Fidelity Outside of Task-Oriented Dialogue}
\label{app:generalization}

\begin{table}[t]
\centering
\footnotesize
\begin{tabular}{llc@{ / }cc@{ / }ccccc}
\toprule
\multirow{2}{*}{\textbf{Base}} & \multicolumn{1}{l|}{\multirow{2}{*}{\textbf{Model}}} & \multicolumn{4}{c|}{\textbf{Naturalness}} & \multicolumn{2}{c|}{\textbf{Mimicry}} & \multicolumn{2}{c}{\textbf{Coverage}} \\
 & \multicolumn{1}{l|}{} & \multicolumn{2}{c}{S2R-N ↑} & \multicolumn{2}{c|}{TT ↓} & AVA ↑ & \multicolumn{1}{c|}{PT3 ↑} & S2R-C ↑ & SD-C ↑ \\ \midrule
\multicolumn{10}{l}{\textbf{SimulatorArena}} \\ \midrule
\multirow{4}{*}{\begin{tabular}[c]{@{}l@{}}Llama \\ 3.1 8B\end{tabular}} & \multicolumn{1}{l|}{UserLM} & -- & \textbf{0.64} & -- & \multicolumn{1}{c|}{0.37} & -- & \multicolumn{1}{c|}{} & \textbf{0.84} & \textbf{0.62} \\
 & \multicolumn{1}{l|}{USP} & 0.41 & 0.38 & 0.40 & \multicolumn{1}{c|}{0.42} & \textbf{0.50} & \multicolumn{1}{c|}{0.01} & 0.11 & 0.52 \\
 & \multicolumn{1}{l|}{RealUserSim} & \textbf{0.29} & 0.21 & 0.19 & \multicolumn{1}{c|}{0.39} & \textbf{0.50} & \multicolumn{1}{c|}{0.05} & 0.00 & 0.50 \\
 & \multicolumn{1}{l|}{CUE} & 0.23 & 0.27 & \textbf{0.17} & \multicolumn{1}{c|}{\textbf{0.21}} & 0.44 & \multicolumn{1}{c|}{\textbf{0.06}} & \textbf{0.83} & 0.56 \\ \midrule
\multirow{2}{*}{\begin{tabular}[c]{@{}l@{}}GPT 5.4 \\ Mini\end{tabular}} & \multicolumn{1}{l|}{RealUserSim} & \textbf{0.42} & 0.27 & 0.12 & \multicolumn{1}{c|}{\textbf{0.12}} & \textbf{0.43} & \multicolumn{1}{c|}{0.09} & 0.45 & \textbf{0.50} \\
 & \multicolumn{1}{l|}{CUE} & \textbf{0.42} & \textbf{0.47} & \textbf{0.13} & \multicolumn{1}{c|}{0.15} & 0.33 & \multicolumn{1}{c|}{\textbf{0.12}} & \textbf{0.87} & \textbf{0.50} \\ \midrule
\multirow{2}{*}{\begin{tabular}[c]{@{}l@{}}Gemini 3.5 \\ Flash Lite\end{tabular}} & \multicolumn{1}{l|}{RealUserSim} & \textbf{0.77} & 0.63 & 0.10 & \multicolumn{1}{c|}{0.20} & \textbf{0.51} & \multicolumn{1}{c|}{\textbf{0.15}} & 0.24 & 0.55 \\
 & \multicolumn{1}{l|}{CUE} & \textbf{0.77} & \textbf{0.80} & \textbf{0.06} & \multicolumn{1}{c|}{\textbf{0.09}} & 0.42 & \multicolumn{1}{c|}{\textbf{0.15}} & \textbf{0.90} & \textbf{0.60} \\ \midrule
\multicolumn{10}{l}{\textbf{PRISM}} \\ \midrule
\multirow{4}{*}{\begin{tabular}[c]{@{}l@{}}Llama \\ 3.1 8B\end{tabular}} & \multicolumn{1}{l|}{UserLM} & -- & 0.72 & -- & \multicolumn{1}{c|}{0.31} & -- & \multicolumn{1}{c|}{} & \textbf{0.95} & \textbf{0.75} \\
 & \multicolumn{1}{l|}{USP} & \textbf{0.82} & \textbf{0.82} & 0.17 & \multicolumn{1}{c|}{0.19} & \textbf{0.56} & \multicolumn{1}{c|}{0.25} & 0.86 & \textbf{0.75} \\
 & \multicolumn{1}{l|}{RealUserSim} & 0.46 & 0.46 & \textbf{0.05} & \multicolumn{1}{c|}{\textbf{0.17}} & 0.54 & \multicolumn{1}{c|}{0.21} & 0.87 & 0.70 \\
 & \multicolumn{1}{l|}{CUE} & 0.56 & 0.70 & 0.15 & \multicolumn{1}{c|}{0.19} & 0.52 & \multicolumn{1}{c|}{\textbf{0.38}} & \textbf{0.95} & \textbf{0.75} \\ \midrule
\multirow{2}{*}{\begin{tabular}[c]{@{}l@{}}GPT 5.4 \\ Mini\end{tabular}} & \multicolumn{1}{l|}{RealUserSim} & 0.51 & 0.38 & 0.10 & \multicolumn{1}{c|}{0.18} & 0.43 & \multicolumn{1}{c|}{0.28} & 0.90 & 0.68 \\
 & \multicolumn{1}{l|}{CUE} & \textbf{0.61} & \textbf{0.67} & \textbf{0.02} & \multicolumn{1}{c|}{\textbf{0.13}} & \textbf{0.49} & \multicolumn{1}{c|}{\textbf{0.43}} & \textbf{0.96} & \textbf{0.72} \\ \midrule
\multirow{2}{*}{\begin{tabular}[c]{@{}l@{}}Gemini 3.5 \\ Flash Lite\end{tabular}} & \multicolumn{1}{l|}{RealUserSim} & \textbf{0.86} & 0.58 & 0.15 & \multicolumn{1}{c|}{0.21} & 0.48 & \multicolumn{1}{c|}{0.35} & 0.91 & 0.72 \\
 & \multicolumn{1}{l|}{CUE} & 0.83 & \textbf{0.89} & \textbf{0.06} & \multicolumn{1}{c|}{\textbf{0.12}} & \textbf{0.59} & \multicolumn{1}{c|}{\textbf{0.51}} & \textbf{0.96} & \textbf{0.77} \\ \bottomrule
\end{tabular}

\caption{\textbf{Out-of-domain user fidelity results for SimulatorArena and PRISM.} Naturalness results are reported for user-conditioned/sampled methods while mimicry is only reported for user-conditioned methods and coverage is only reported for sampled methods.}
\label{tab:out_fidelity}
\end{table}

Table~\ref{tab:out_fidelity} reports all six fidelity metrics on SimulatorArena \citep{dou2025simulatorarena} and PRISM \citep{kirk2024prism}.
Section~\ref{sec:generalization} summarizes the main cross-domain findings; here, we focus on what the full metric grid reveals about how the fidelity measures behave outside task-oriented dialogue.

\paragraph{Naturalness generally improves with stronger simulator bases.}
For CUE, naturalness improves with simulator-base capability across both out-of-domain datasets, although not every metric is strictly monotonic.
On SimulatorArena, sampled S2R-N increases from 0.27 $\rightarrow$ 0.47 $\rightarrow$ 0.80 and TT decreases from 0.21 $\rightarrow$ 0.15 $\rightarrow$ 0.09 across Llama 3.1 8B, GPT-5.4 Mini, and Gemini 3.5 Flash Lite.
On PRISM, the corresponding values are 0.70 $\rightarrow$ 0.67 $\rightarrow$ 0.89 for S2R-N and 0.19 $\rightarrow$ 0.13 $\rightarrow$ 0.12 for TT.
RealUserSim exhibits a similar overall trend at lower naturalness levels.
Together, these results support the observation in Section~\ref{sec:cueornot} that stronger simulator bases need not reduce user realism when persona conditioning steers them away from assistant-like defaults, contrasting with the inverse scaling behavior reported for unconditioned user simulators by \citet{naous2025flippingthedialogue}.

\paragraph{Mimicry exhibits a strong domain effect.}
PT3 scores are uniformly low on SimulatorArena: the best cell is 0.15 and the worst is 0.01, compared with 0.21--0.51 on PRISM.
We interpret this primarily as a property of the evaluation domain.
SimulatorArena interactions are dominated by short, task-focused writing and mathematics requests, leaving relatively little user-specific behavioral or stylistic evidence for a paired judge to recover.
Mimicry metrics should therefore be interpreted cautiously in collaborative task domains: a simulator may have lower observed outcome error (Table~\ref{tab:simarenamae}) even when its trajectory contains too little identifying signal to match a particular human user.

\paragraph{AVA is weakly discriminative out of domain.}
Across the fourteen reported out-of-domain AVA cells, AVA ranges from 0.33 to 0.59 and frequently produces orderings inconsistent with PT3.
On SimulatorArena, for example, CUE scores below RealUserSim on AVA at every base (0.44/0.33/0.42 versus 0.50/0.43/0.51), while matching or exceeding it on PT3.
This pattern is consistent with the paired-versus-shuffled ablation in Table~\ref{tab:cuemimicry}, where replacing an episode's CUE with another user's representation changes AVA only from 0.27 to 0.20.
A metric that responds only weakly to whether the correct user is paired with the trajectory provides limited resolution for user-specific mimicry.
We therefore retain AVA for comparability with prior work \citep{wang2025knowyoufirst}, but place greater weight on PT3 when interpreting paired-session fidelity.

\paragraph{S2R-C discriminates on task-like domains but saturates on casual conversation.}
On SimulatorArena, S2R-C separates methods substantially: USP and RealUserSim span 0.00--0.45, whereas CUE reaches 0.83--0.90 and UserLM reaches 0.84.
This reproduces the sampling-mechanism differences discussed in Section~\ref{sec:fidelity}.
On PRISM, however, all methods score between 0.86 and 0.96, leaving little separation.
This behavior is consistent with the provenance of the underlying behavioral taxonomy: the Sim2Real gaps characterized by \citet{zhou2026sim2real} were derived from task-oriented interactions, so the resulting feature space is less discriminative on open-ended conversational data.
RealUserSim's 0.00 score on SimulatorArena with Llama 3.1 8B is additionally a consequence of clipping in the normalized Chamfer definition rather than evidence of a qualitatively distinct failure mode.

\paragraph{SD-C compresses differences visible at finer scales.}
SD-C varies within a relatively narrow range across both methods and simulator bases: 0.50--0.62 on SimulatorArena and 0.68--0.77 on PRISM.
On PRISM, Figure~\ref{fig:prismsweep} shows that a finer-grained neighborhood analysis reveals substantially larger differences.
At radius 0.05, the fraction of sampled users with at least one real-user neighbor ranges from 67\% to 81\% across methods and diverges further as the radius tightens, while SD-C separates the same methods by less than 0.10.

This compression follows from the bidirectional nearest-neighbor aggregation used by Chamfer distance.
A simulated population can achieve good aggregate coverage while occupying intermediate regions between human modes, even if comparatively few generated users closely match individual human neighborhoods.
We therefore report the radius sweep alongside SD-C to expose local coverage structure that is obscured by the aggregate Chamfer score.

\paragraph{Domain match can resemble generalization.}
USP's S2R-N scores across the three domains are 0.26/0.27 on $\tau^2$-Bench, 0.41/0.38 on SimulatorArena, and 0.82/0.82 on PRISM.
This ordering also tracks similarity to LMSYS-Chat, the primary corpus used to train USP \citep{wang2025knowyoufirst,zheng2024lmsys}: among the three evaluation settings, PRISM most closely resembles open-ended conversational interaction.
Its strong PRISM performance should therefore not be interpreted straightforwardly as evidence of domain-independent generalization.
An alternative reading is that USP performs best where the evaluation distribution most resembles its training data and degrades as the interaction setting moves farther from that distribution.

\clearpage
\section{Single-Task Case Study: Reproducing a Human Interaction's Constraint Loss}
\label{app:case_study}
This case follows a human interaction, its logged CUE description, and two simulated replays of \texttt{retail\_49\_ann2} (native retail task 49). We compare paired CUE with None, the no-persona baseline, using GPT-5.4-mini user simulators and a GPT-5.2 customer-service agent in seed~0. The human reference also uses GPT-5.2. Both the human and CUE interactions execute the same incorrect exchange, whereas None selects the correct replacement. The match is in the concrete tool action, not only the binary failure outcome.
This case follows a human interaction, its logged CUE description, and simulated replays of \texttt{retail\_49\_ann2} (native retail task 49). We compare paired CUE with None, the no-persona baseline, using GPT-5.4-mini user simulators and a GPT-5.2 customer-service agent in seed~0. The human reference also uses GPT-5.2. Both the human and CUE interactions execute the same incorrect exchange, whereas None selects the correct replacement. The match is in the concrete tool action, not only the binary failure outcome.

\subsection{Task and human reference}
The customer mistakenly ordered an IPX7 wireless earbud and wants to exchange it for the cheapest earbud among the other items in the same order, matching their IPX4 water resistance. The human and simulator task instructions both specify ``the cheapest earbud item from the rest of that order.'' Thus, the candidate set is restricted to earbuds already in the order, rather than all available variants in the product catalog.

Order \texttt{\#W3470184} contains the original IPX7 item \texttt{2757705742} (\$258.97) and two IPX4 earbuds. The cheaper of those two, \texttt{1646531091} (\$232.49), is the required replacement. A cheaper IPX4 variant, \texttt{8555936349} (\$226.49), exists in the catalog but is absent from the order. Selecting it satisfies the water-resistance preference while violating the within-order restriction.

\begin{fmadialogue}{Human interaction: the within-order restriction is lost}

The following are selected verbatim dialogue excerpts; intervening turns and tool lookups are omitted.
\fmaturn{Human}{Sorry, I meant you should change it to IPX4 and pick the cheapest option.}
\fmaturn{Agent}{Got it --- change from \textbf{IPX7 $\rightarrow$ IPX4}, and pick the \textbf{cheapest available IPX4} option.}
\fmaturn{Agent proposal (summary)}{Exchange item 2757705742 for catalog variant 8555936349, with a \$32.48 refund to the original gift card.}
\fmaturn{Human}{yes}
\fmaturn{Recorded action}{The agent executes the exchange to 8555936349. Recorded reward: 0.}
\end{fmadialogue}

\paragraph{What the replay should recover}
The relevant failure is the loss of a task constraint during an otherwise plausible exchange. The human approves the catalog-wide cheapest option, and the agent carries out that choice. Reproducing this interaction requires more than generating an arbitrary failed episode: the simulation must arrive at the same incorrect replacement despite receiving the original within-order restriction.

\clearpage
\subsection{Complete logged CUE instruction}
\label{app:case_cue_instruction}
The following is the complete \texttt{command\_block} logged with the paired CUE rollout, including all behavioral commands, eight retrieved style sketches, and writing-style commands. Wording and sample punctuation are preserved; headings and bullets are typeset for readability. The task instructions are supplied separately.

\begin{fmadialogue}{Complete CUE description for \texttt{retail\_49\_ann2}}
\smallskip\textbf{General behavior:}
\begin{itemize}[leftmargin=*,nosep]
\item Keep the tone lightly conversational and human, with small acknowledgments and simple closers rather than polished assistant-like phrasing.
\item Use a brief closing once the issue is resolved, with a simple thanks and goodbye.
\item Stay focused on the immediate task and avoid adding extra context, backstory, or side questions unless the assistant asks for it.
\item Use short, plain, conversational phrasing with minimal punctuation and no polished assistant-like wording.
\end{itemize}
\smallskip\textbf{User-specific behavior:}
\begin{itemize}[leftmargin=*,nosep]
\item If the assistant asks for a missing detail, provide it in a compact follow-up rather than a long explanation.
\item If the assistant asks for a missing detail, supply it in a compact follow-up rather than restarting the whole request.
\item If the assistant gives a partial or mismatched answer, respond by narrowing the request and asking for the exact missing detail.
\item If the assistant asks whether anything else is needed, close the exchange quickly with a brief thanks or no-more-needed reply.
\item If the assistant asks a question that is not the one you were expecting, answer the earlier thread instead of the latest prompt.
\end{itemize}
\smallskip\textbf{Style samples} -- short sketches of how users like this one type. Match their mechanics, not their wording: punctuation (including missing, doubled, or absent end marks), capitalization, spacing, contractions, and how short the messages are. Angle-bracket slots like \textless{}ENTITY\textgreater{} or \textless{}NUM\textgreater{} stand in for details -- put your own specifics there and never write the brackets. Take nothing else from them, including their task content:
\begin{itemize}[leftmargin=*,nosep]
\item ok . thanks . that works .
\item yes . please .
\item that works .
\item thanks . that works .
\item can you help me find a church in [value\_place] ?
\item i am looking for a place to dine that is [restaurant\_pricerange]
\item it s called [attraction\_name] .
\item i am not interested in [restaurant\_food] .
\end{itemize}
\smallskip\textbf{Writing style:}
\begin{itemize}[leftmargin=*,nosep]
\item Use lowercase, casual phrasing with minimal capitalization.
\item Use polite gratitude sparingly, usually as a brief closing.
\item Use short, casual fragments and keep replies very brief.
\item Keep punctuation sparse and informal, often using periods with spaces around them.
\item Use short, direct fragments when giving simple instructions or confirmations.
\end{itemize}

\end{fmadialogue}

The saved decoder field is marked ``precomputed.'' The block contains no mention of earbuds, item identifiers, water resistance, or the within-order restriction, and no instruction to make an incorrect choice. It includes both brief-confirmation style examples and a command to seek clarification when an answer is mismatched.

\clearpage
\subsection{Simulated interaction and executed outcomes}
The excerpts below preserve the simulator's wording. Agent proposals are summarized explicitly. CUE and None interact with the agent in separate closed-loop conversations; the agent's histories and presentation of the alternatives differ.

\begin{fmadialogue}{CUE: states the restriction but accepts the broader choice}
\fmaturn{Opening user turn (excerpt)}{i want to swap it for the cheapest other earbud in the order with the same water resistance as the rest .}
\fmaturn{Agent proposal (summary)}{The cheapest available IPX4 variant is 8555936349 at \$226.49. The cheapest IPX4 earbud already in the order is 1646531091 at \$232.49. Choose a replacement and confirm the payment method and exchange.}
\fmaturn{User}{yes . do the cheapest available ipx4 one , 8555936349 . refund the difference to the original method . please proceed .}
\fmaturn{Recorded action}{The agent exchanges to 8555936349, the same incorrect replacement as in the human interaction.}
\fmaturn{User}{thanks . that works . \#\#\#STOP\#\#\#}

\end{fmadialogue}

\begin{fmadialogue}{None: preserves the within-order restriction}

\fmaturn{Agent proposal (summary)}{Option A is 1646531091, the cheapest IPX4 earbud already in the order. Option B is 8555936349, the cheapest available IPX4 earbud overall, which is not currently in the order.}
\fmaturn{User}{A. Please go ahead and exchange item 2757705742 to item 1646531091. Yes, I confirm this is the only earbud I'm exchanging.}
\fmaturn{Recorded action}{The agent exchanges to 1646531091, satisfying the original restriction.}

\end{fmadialogue}

\begin{table}[h]
\centering\small
\begin{tabular}{llrc}
\toprule
Interaction & Executed replacement & Refund & Reward \\
\midrule
Human & \texttt{8555936349} (outside order) & \$32.48 & 0 \\
CUE, seed 0 & \texttt{8555936349} (outside order) & \$32.48 & 0 \\
None, seed 0 & \texttt{1646531091} (within order) & \$26.48 & 1 \\
\bottomrule
\end{tabular}
\caption{Executed outcomes for the selected episode. All three interactions call \texttt{exchange\_delivered\_order\_items} on order \texttt{\#W3470184}, replacing item \texttt{2757705742} and using \texttt{gift\_card\_7245904}. The native task requires replacement \texttt{1646531091}.}
\label{tab:case_exchange}
\par
\end{table}

\paragraph{Interpretation and seed context}
The CUE replay reproduces the human interaction's specific constraint loss without an explicit instruction to drop the restriction or choose the wrong item. Both users approve the final choice, so this is not an unauthorized action. The match illustrates failure reproduction in a selected interaction; it does not isolate the effect of any individual command or retrieved example. On this task, paired CUE succeeds in GPT seeds~1 and~2, while None succeeds in all three seeds.

\clearpage
\subsection{Additional baseline trajectories on the same episode}
\label{app:case_baselines}
Table~\ref{tab:case_baselines} extends the seed-0 comparison to the available PPol, paired RealUserSim without examples, UserLM, and USP trajectories. Prompt-based methods use GPT-5.4-mini; UserLM and USP use their released trained models. Each entry describes a saved closed-loop interaction, with its own agent history. Matching a failure score does not imply reproducing the human's failure: the other failed runs below never execute an exchange.

\begin{table}[h]
\centering\small
\begin{tabular}{p{0.22\linewidth}cp{0.49\linewidth}c}
\toprule
Interaction & Reward & Executed outcome & Match? \\
\midrule
Human & 0 & Incorrect target \texttt{8555936349} & Ref. \\
CUE paired & 0 & Same incorrect target \texttt{8555936349} & Yes \\
None & 1 & Correct target \texttt{1646531091} & No \\
PPol & 1 & Corrects its choice before execution & No \\
RealUserSim paired, no examples & 0 & No exchange; unsupported order ID & No \\
UserLM & 0 & No exchange; unsupported item IDs & No \\
USP paired & 0 & No exchange; authentication fails & No \\
USP sampled & 0 & No exchange; off-task dialogue & No \\
\bottomrule
\end{tabular}
\caption{Observed outcomes on the selected episode and seed. ``Match'' requires the same incorrect exchange as the human, not merely reward 0. USP sampled is a population-sampling reference, not paired replay. Prompt provenance limits are described below.}
\label{tab:case_baselines}
\end{table}

\begin{fmadialogue}{PPol: initially chooses the wrong item but repairs the request}
\fmaturn{User}{i want item 2757705742 swapped to the cheapest available ipx4 earbud, and use the gift card for any difference.}
\fmaturn{Agent proposal (summary)}{Confirm exchange to 8555936349, with a \$32.48 refund to the original gift card.}
\fmaturn{User}{no, i'd rather exchange it to the cheapest earbud already in the order and keep the gift card out of it}
\fmaturn{Agent response (summary)}{Offers the correct target 1646531091 and explains that the refund must use the original payment method or an existing gift card.}
\fmaturn{User}{yes}
\fmaturn{Recorded action}{Exchanges to 1646531091. Reward: 1.}
\end{fmadialogue}

PPol's saved trajectory initially follows the same catalog-wide choice as the human and CUE, but restores the within-order restriction before execution. However, the persona record associated with this episode describes a different customer seeking an airline refund. We reproduce that record and its full prompt wrapper below rather than substituting a plausible retail persona. This mismatch prevents treating the trajectory as a verified comparison under task-appropriate persona conditioning.

\clearpage
\begin{fmadialogue}{RealUserSim paired without examples: unsupported order ID}
\fmaturn{User}{Sure, here's the order ID: \#W4839216. Please check it as soon as you can.}
\fmaturn{Agent response (summary)}{The lookup returns ``Order not found''; the agent asks the user to check the identifier.}
\fmaturn{User}{i only have that order id \#\#\#STOP\#\#\#}
\fmaturn{Recorded outcome}{No exchange. Reward: 0.}
\end{fmadialogue}

\begin{fmadialogue}{UserLM and USP: failures before an exchange}
\fmaturn{UserLM opening}{Hi! How can I help you today?}
\fmaturn{Subsequent behavior (summary)}{UserLM supplies the correct name and ZIP, then an unsupported item ID and a long repetition of another identifier. It stops after the agent asks it to identify a valid item.}
\fmaturn{USP paired identification turn}{My fist name NAME\_1, Lastname NAME\_2, zipcode 19121, i want retun}
\fmaturn{Subsequent behavior (summary)}{The agent cannot authenticate the supplied identity. USP then emits a structured customer object and stops. The sampled USP run instead drifts into personal narratives, mathematics, and unrelated requests.}
\fmaturn{Recorded outcomes}{All three runs receive reward 0 without an exchange.}
\end{fmadialogue}

\end{document}